\documentclass[a4paper,fleqn]{cas-sc}

\usepackage[authoryear,longnamesfirst]{natbib}
\usepackage{booktabs}       % professional-quality tables
\usepackage{amsfonts}       % blackboard math symbols
\usepackage{nicefrac}       % compact symbols for 1/2, etc.
\usepackage{microtype}      % microtypography
\usepackage{xcolor}         % colors
\usepackage{bbding}
\usepackage{subfigure}
\usepackage{booktabs} % for professional tables
\usepackage{amsmath}
\usepackage{float}
\usepackage{amssymb}
\usepackage{enumerate}
\usepackage{color}
\usepackage{mathtools}
\usepackage{multirow}
\usepackage{caption}
\usepackage{algorithm}
\usepackage{algorithmicx}
\usepackage{algpseudocode}
\usepackage{wrapfig}
\floatname{algorithm}{Algorithm} %算法
\newcommand*{\imgg}[1]{%
    \raisebox{-.02\baselineskip}{%
        \includegraphics[
        height=\baselineskip,
        width=\baselineskip,
        keepaspectratio,
        ]{#1}%
    }%
}
\def\tsc#1{\csdef{#1}{\textsc{\lowercase{#1}}\xspace}}
\tsc{WGM}
\tsc{QE}
\begin{document}
\let\WriteBookmarks\relax
\def\floatpagepagefraction{1}
\def\textpagefraction{.001}

\shorttitle{A Two-stage Reinforcement Learning-based Approach for Multi-entity Task Allocation}    

\shortauthors{A. Gong, K. Yang et al.}  

\title [mode = title]{A Two-stage Reinforcement Learning-based Approach for Multi-entity Task Allocation}  

\tnotemark[1] 

\tnotetext[1]{This work was supported in part by the Science and Technology Innovation 2030-Key Project under Grant 2021ZD0201404.} 

% \tnotemark[<tnote number>] 


% \author[1,3]{Author Name}[type=editor,

\author[1]{Aicheng Gong}

% \cormark[1]

\fnmark[1]

\ead{gac19@mails.tsinghua.edu.cn}

% \ead[url]{<URL>}

% \credit{<Credit authorship details>}

% \affiliation[1]{organization={China Nuclear Power Engineering Company Ltd.},
%             city={Shenzhen},
% %          citysep={}, % Uncomment if no comma needed between city and postcode
%             postcode={518172}, 
%             country={China}}

\author[1]{Kai Yang}

% Footnote of the second author
\fnmark[1]

% Email id of the second author
\ead{yk22@mails.tsinghua.edu.cn}


\affiliation[1]{organization={Shenzhen International Graduate School, Tsinghua University},
            city={Shenzhen},
%          citysep={}, % Uncomment if no comma needed between city and postcode
            postcode={518055}, 
            country={China}}
\author[1]{Jiafei Lyu}

% Footnote of the second author
% \fnmark[3]

% Email id of the second author
\ead{lvjf20@mails.tsinghua.edu.cn}



\author[1]{Xiu Li}[orcid=0000-0003-0403-1923]

\fnmark[$^*$]

% Footnote of the second author
% \fnmark[4]

% Email id of the second author
\ead{li.xiu@sz.tsinghua.edu.cn}

\cortext[1]{Corresponding author}
\fntext[1]{The first two authors contribute equally to this work.}



% !!!!!!!这里想要变成蓝色 但是好像有问题  页数角标也有问题 显示24/23(fixed)
\begin{abstract}
Task allocation is a key combinatorial optimization problem, crucial for modern applications such as multi-robot cooperation and resource scheduling. Decision makers must allocate entities to tasks reasonably across different scenarios. However, traditional methods assume static attributes and numbers of tasks and entities, often relying on dynamic programming and heuristic algorithms for solutions. In reality, task allocation resembles Markov decision processes, with dynamically changing task and entity attributes. Thus, algorithms must dynamically allocate tasks based on their states. To address this issue, we propose a two-stage task allocation algorithm based on similarity, utilizing reinforcement learning to learn allocation strategies. The proposed pre-assign strategy allows entities to preselect appropriate tasks, effectively avoiding local optima and thereby better finding the optimal allocation. We also introduce an attention mechanism and a hyperparameter network structure to adapt to the changing number and attributes of entities and tasks, enabling our network structure to generalize to new tasks. Experimental results across multiple environments demonstrate that our algorithm effectively addresses the challenges of dynamic task allocation in practical applications. Compared to heuristic algorithms like genetic algorithms, our reinforcement learning approach better solves dynamic allocation problems and achieves zero-shot generalization to new tasks with good performance. The code is available at \href{https://github.com/yk7333/TaskAllocation}{https://github.com/yk7333/TaskAllocation.}
\end{abstract}


\maketitle

% Main text
\section{Introduction}
Task allocation is a critical combinatorial optimization problem (\cite{yan2012multi}). It plays a vital role in modern applications such as multi-robot collaboration, resource scheduling, and more. In warehouse logistics, mobile robot teams allocate tasks and transport goods to different locations (\cite{2021A}). In industrial and manufacturing fields, robotic arms are assigned to various processing tasks (\cite{johannsmeier2016hierarchical}). On-demand carpooling and delivery services require the dispatch of agents to meet customer needs (\cite{hyland2018dynamic}). As decision-makers, we must allocate tasks to entities based on their location, current attributes, and task requirements to ensure timely task completion.

\textcolor{black}{Various approaches have been proposed to solve the task allocation problem (\cite{wang2023survey,quinton2023market,burkard2012assignment,parasuraman1996effects,alighanbari2005decentralized,merlo2023ergonomic}). The common approach treats it as a discrete optimization problem, assigning entities to tasks. Accurate solution algorithms like the Hungarian algorithm (\cite{msala2023new,samiei2023distributed,samiei2020distributed,munkres1957algorithms}), branch and bound (\cite{martin2021multi,singh2021scheduling,lawler1966branch}), network flow algorithms (\cite{javanmardi2023s,jamil2023irats,de2007distributed}), fuzzy logic algorithm (\cite{ali2024mobility,sharma2023optimized,ali2021real,jamil2022resource}) and dynamic programming (\cite{choudhury2022dynamic,alkaabneh2021unified,bellman2010dynamic}) are used, as well as heuristic algorithms such as the genetic algorithm (\cite{deng2023multi,patel2020decentralized,ye2020cooperative,page2010multi,forrest1996genetic,fu2023task}), particle swarm optimization (\cite{kalimuthu2024design,geng2021particle,qingtian2021application}), symbiotic organisms search(\cite{truong2020quasi,gharehchopogh2020comprehensive,abdullahi2023adaptive}). simulated annealing (\cite{barbosa2023optimization,wang2022research,barboza2024task,bertsimas1993simulated}) are used to solve this problem. There are also methods using game theory to allocate robots to finish tasks \cite{martin2023multi,sun2023bargain}. 
However, while these methods have proven effective in addressing the task allocation problem under static conditions, they often assume static task and entity attributes. In reality, task assignments frequently encounter dynamic states and fluctuations in task estimation or entity numbers over time. As a result, the applicability of the aforementioned methods may diminish, necessitating the exploration of new approaches for dynamic task allocation. This dynamic nature underscores the importance of developing methodologies capable of adapting to evolving conditions and uncertainties in real-time task assignment scenarios.}

To address the dynamic allocation problem in real-world scenarios, our aim is to develop more efficient and practical algorithms compared to prior methods. After each step, a series of tasks can emerge, requiring us to allocate entities effectively to achieve task goals. However, these resources come with associated costs, such as power consumption in transmission and additional expenses incurred by companies when assigning employees to extra tasks. Therefore, we must carefully select the appropriate resources from a large pool to minimize costs and successfully complete tasks. 
Treating this problem as a Markov Decision Process (MDP), we leverage Reinforcement Learning (RL) algorithms to dynamically adjust allocation actions based on the current state, without incurring additional computational costs. \textcolor{black}{Several RL-based methods have been developed for task allocation in various domains. For instance, \cite{afrin2023dynamic} integrates edge and cloud computing with robotics to address computation tasks impacted by uncertain failures, utilizing a multiple deep Q-network mechanism for dynamic task allocation and ensuring quicker resiliency management. Similarly, \cite{fu2023dynamic} applies deep reinforcement learning to mobile crowd sensing, using a double deep Q network based on the dueling architecture to manage dynamic task allocation under multiple constraints, surpassing traditional heuristic methods in platform profit and task completion rate. These approaches show promising results in Smart Farm and mobile crowd sensing scenarios. Additionally, methods such as \cite{park2021cooperative} and \cite{agrawal2023rtaw} use attention modules to extract task and robot properties, facilitating efficient allocation decisions in environments like warehouses. Other notable works include task allocation for workers in spatio-temporal crowdsourcing (\cite{zhao2023task}) and mobile crowdsensing (\cite{xu2023intelligent}).}

While RL algorithms have been widely used, there is a lack of research on allocating varying numbers of agents and handling scenarios with emerging tasks. Simply adopting traditional RL algorithms is inadequate for handling dynamically changing numbers and attributes of agents in practical applications. \textcolor{black}{This is because RL predefines the number of agents and establishes a fixed network structure to calculate the value of each agent's actions in each state. When the number of agents suddenly increases, the neural network cannot allocate tasks for the additional agents.} The methods mentioned above, which utilize reinforcement learning, only consider problems with a fixed number of agents and aim to improve the coordination of existing robots to accomplish tasks more effectively. Additionally, these robots have fixed characteristics, and there is no need to consider issues such as which agents' resource attributes are more suitable for allocation or which agents remain stationary on standby. Hence, to address this challenge, this paper introduces a two-stage task allocation algorithm that leverages the similarity between agent and task attributes, rendering RL algorithms applicable to the task allocation problem. \textcolor{black}{Compared with heuristic algorithms, our proposed reinforcement learning method can achieve better results in dynamic task allocation and easily solve the problem of variable number of entities. When the attributes or number of tasks or entities change, our method can achieve good generalization results without training, and can better handle a series of allocation problems.} Our main contributions are fourfold:
\begin{enumerate}
\item We introduce a pre-assign method for efficient task allocation. Firstly, the agent is pre-assigned to a task as a candidate agent for task selection, and then the agent is selected based on the degree of correlation between the task and the agent.
\item We incorporate Actor-Critic structure in combinational optimization problems by proposing a two-head attention mechanism module. It calculates the values of Actor and Critic simultaneously based on the similarity between agent and task attributes, enabling task allocation for a variable number of agents and zero-shot generalization of new tasks.
\item An attention-based hyperparameter network structure is proposed to estimate the overall value of critical outputs for different numbers of agents, facilitating fine-tuning of the variable number of agents in new scenarios.
\item We propose a seq2seq-like structure, similar to PointNet, to select pre-assigned agents. It selects an appropriate number of agents with suitable attributes for each task.
\end{enumerate}
\textcolor{black}{For your convenience, please refer to Table \ref{ab}, which includes a comprehensive list of all the abbreviations used in the article.}

\begin{table}[h]
    \centering
    \resizebox{0.75\linewidth}{55mm}{%
    \begin{tabular}{ll}
        \toprule
        \textcolor{black}{\textbf{Abbreviation}} & \textcolor{black}{\textbf{Definition}} \\
        \midrule
        \textcolor{black}{RL} & \textcolor{black}{Reinforcement Learning} \\
        \textcolor{black}{NP} & \textcolor{black}{Non-deterministic Polynomial} \\
        \textcolor{black}{MDP} & \textcolor{black}{Markov Decision Process} \\
        \textcolor{black}{UAV} & \textcolor{black}{Unmanned Aerial Vehicle} \\
        \textcolor{black}{AMIX} & \textcolor{black}{Attention MIX network} \\
        \textcolor{black}{SHN} & \textcolor{black}{Self-attention-based Hyper Network} \\
        \textcolor{black}{SAC} & \textcolor{black}{Soft Actor-Critic method} \\
        \textcolor{black}{QMIX} & \textcolor{black}{Q MIXing method} \\
        \textcolor{black}{DDPG} & \textcolor{black}{Deep Deterministic Policy Gradient method} \\
        \textcolor{black}{EPT} & \textcolor{black}{Electric Power Transportation environment} \\
        \textcolor{black}{LBF} & \textcolor{black}{Level-Based Foraging environment} \\
        \textcolor{black}{RBF} & \textcolor{black}{Resource-Based Foraging environment} \\
        \textcolor{black}{MT} & \textcolor{black}{Material Transportation environment} \\
        \textcolor{black}{DVRP} & \textcolor{black}{Dynamic Vehicle Routing Problem} \\
        \textcolor{black}{GA} & \textcolor{black}{Genetic Algorithm} \\
        \textcolor{black}{PSO} & \textcolor{black}{Particle Swarm Optimization} \\
        \textcolor{black}{SOS} & \textcolor{black}{Symbiotic Organisms Search} \\
        \bottomrule
    \end{tabular}%
    }
    \caption{\textcolor{black}{Table of abbreviations}}
    \label{ab}
\end{table}

\section{Related Work}
\textbf{Multi-robot task allocation}: The overview of task allocation for multiple robots is how to allocate subtasks to each robot in a way that balances the entire load. Due to the \textcolor{black}{Non-deterministic Polynomial (NP)} difficulty of the problem, researchers decompose it into sub-problems or apply meta-heuristic techniques. Some studies have addressed the challenges in multi-robot task allocation, which have various objectives such as energy consumption, time, cost, fairness, and task completion. Genetic algorithm (\cite{patel2020decentralized,ye2020cooperative,page2010multi}), k-means clustering algorithm (\cite{muthusamy2021cluster,sheikh2023machine,elango2011balancing}), and imitative learning (\cite{wang2020imitation, yuvaraj2021improved, jebara2001discriminative}) are used to solve this problem. \textcolor{black}{Most research has focused on specific applications, such as manufacturing (\cite{wang2021digital,morariu2020machine,giordani2010distributed}), inspection (\cite{karami2020task, zhou2023sliding,liu2017dynamic}), warehouses (\cite{tsang2018novel, albert2023trends}), disaster rescue (\cite{ghassemi2022multi,xu2023discrete}), multiple unmanned aerial vehicle (UAV) formations (\cite{wu2023heuristic, zhang2020cache}), computation allocation (\cite{sun2023joint}), and workshop scheduling (\cite{wang2023ant,zhang2022dynamic,han2022digital,dahl2009multi}).} \textcolor{black}{Currently, (\cite{zhang2023energy}) uses graph-based method to batch tasks with time and precedence constraints, aiming to optimize coordination, reduce downtime, and minimize energy consumption to addresses multi-robot task allocation. (\cite{wen2024indicator}) introduces a unified model for the multi-robot task allocation problem, presenting an efficient indicator-based multi-objective evolutionary algorithm with a hybrid encoding scheme and adaptive archive update mechanism. (\cite{zhang2024scalable}) proposes a novel graph deep-reinforcement-learning-based approach that leverages graph sampling and cross-attention decoding to efficiently allocate tasks to robots, demonstrating high scalability and robustness through extensive experiments in various multi-robot task allocation scenarios. (\cite{yan2023solving}) introducing a hyper-heuristic algorithm that minimizes the time cost for compulsory tasks while selectively completing functional tasks to address multi-robot task allocation. (\cite{guo2024effective}) introducing a collaborative discrete bee colony algorithm. (\cite{agrawal2023rtaw}) presents a novel reinforcement learning-based algorithm (RTAW) for multi-robot task allocation in warehouse environments, utilizing a deep multi-agent reinforcement learning method with an attention-inspired policy architecture to minimize total travel delay and improve makespan.} This article focuses more on algorithmic innovation and provides a new allocation model based on RL algorithms for tasks, including multi-robot allocation. 

\textbf{Dynamic task allocation}: There are several typical methods in the study of dynamic task allocation. The first method is to use a multi-robot task allocation strategy (\cite{quinton2023market,schneider2017mechanism}), which mimics the process of market trading. When a new task is born, all robots will quote, and ultimately the task is assigned to the robot with the lowest quote. Robots will regularly update their quotes to reassign tasks, but this auction is very time-consuming, so the action distance of robots assigned using this strategy will be set very short (\cite{ullah2023improved,de2022decentral,talebpour2018multi}). The second method is to still use the static task allocation method to search for solutions. When the task state changes, fine-tuning methods are used to find new solutions. For example, the genetic algorithm in meta heuristic algorithms can partially cross mutate on the basis of the original solution to obtain a new solution (\cite{patel2020decentralized,zhang2021multitask,chen2018cluster,zhou2019balanced}). Some methods need to rerun the algorithm to calculate new optimal solutions, such as Integer programming algorithm (\cite{chen2023scheduling,li2017research,su2018two}) and search algorithm (\cite{sanaj2020nature,zhang2020dynamic,mitiche2019iterated}). Whether it is fine-tuning or searching for the optimal solution again, it is a very time-consuming process, and when the task changes frequently, it is completely impossible to find a suitable solution. The third method is to use clustering algorithms to cluster similar entities into a formation, and entities with similar functions or attributes will be assigned to a formation, greatly reducing the allocation time (\cite{sun2021and, sarkar2018scalable}). This approach can meet the requirements of real-time performance, but once the attributes of the entity are misvalued or malfunctions occur, the completion rate of the task will be greatly reduced (\cite{mitiche2019iterated}). \textcolor{black}{Currently, Dynamic Multi-Robot Task Allocation for capacitated robots using Satisfiability Modulo Theories solves the problem of handling dynamic streams of tasks with deadlines, ensuring soundness, completeness, and generality for various task specifications \cite{tuck2024smt}. \cite{merlo2023ergonomic} introduces an ergonomic role allocation framework for human-robot collaboration, integrating task features and human state measurements to optimize task assignment and reduce the risk of work-related musculoskeletal disorders. \cite{afrin2023dynamic} incorporates edge and cloud computing with robotics to support computation tasks affected by uncertain failures, proposing a multiple deep Q-network dynamic task allocation mechanism to ensure faster resiliency management. \cite{fu2023dynamic} leverages deep reinforcement learning methods for mobile crowd sensing task allocation, using a double deep Q network based on the dueling architecture to address dynamic task allocation under multiple constraints. These two approaches have shown promising results utilizing deep reinforcement learning in Smart Farm and mobile crowd sensing scenarios.} \textcolor{black}{Existing RL methods (\cite{park2021cooperative, agrawal2023rtaw}) aim to solve the allocation problem by using attention modules to extract task and robot properties for decision-making, similar to our approach. They apply an attention-based approach to derive robot and task embeddings and utilize algorithms to allocate robots, often in warehouse environments. However, these methods model the problem as an MDP and apply reinforcement learning but don't handle scenarios with a variable number of robots. Additionally, they assume fixed robot attributes and do not allow robots to autonomously propose bids. Also, all robots must be allocated, making it a fixed-number allocation problem. In contrast, our approach addresses more complex scenarios. Our entities can include robots, vehicles, suppliers who propose bids based on their attributes, or employees requesting salaries. Our problem definition allows for entity-specific attributes and bids, enabling managers to exclude entities with high bids and low attributes. Our method accommodates variable numbers of entities, dynamic bids, and can handle entity selection and task allocation even when the number of entities far exceeds the number of tasks.}

\begin{figure*}[t]
 \begin{center}
  \includegraphics[height=240pt,width=460pt]{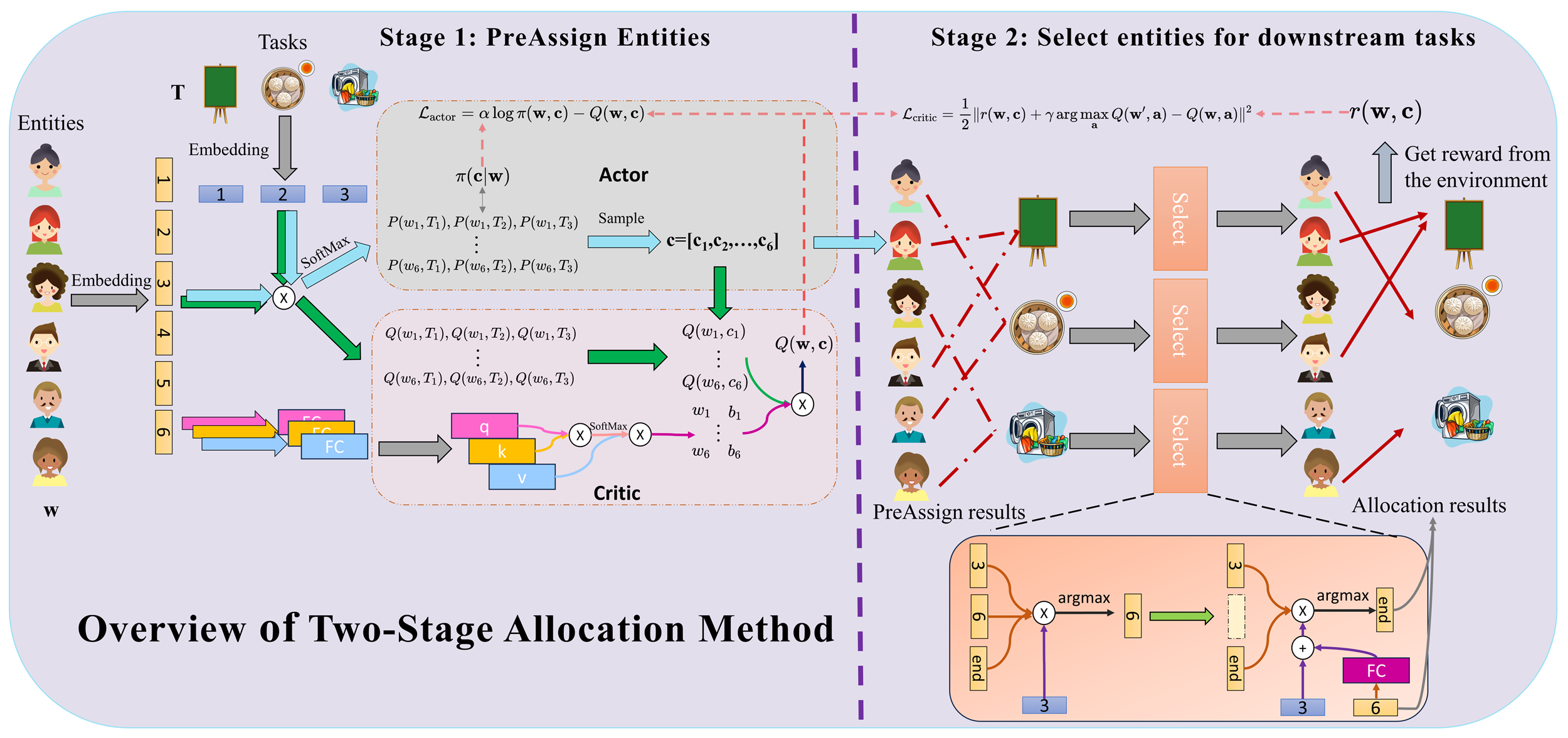}
  \caption{\textcolor{black}{The overview of our Two-Stage allocation method. First, tasks are pre-assigned based on the attributes of both tasks and entities. We utilize the PreAssign model to identify potential entities for each task, and then select the entities best suited to perform each task. Following the pre-assignment phase, each task is matched with suitable entities for its completion, resulting in the final allocation for each task.}}
  \label{overview}
   \end{center}
\end{figure*}

\section{Problem Setup}
In the environment, there are many tasks in an episode, and we need to act as managers to assign entities to solve these tasks in order to receive rewards. These entities can be workers who have their own minds and are self-interested or cars that contain a lot of resources. Choosing these entities to complete tasks requires a cost, such as overtime wages for employees and fuel consumption considerations for allocating vehicles. What we need to do is choosing these entities in a reasonable manner and allocating them to complete various tasks. Suppose there are $m$ tasks and $n$ entities in an environment. The number of entities $n$ is much larger than $m$ and these entities will follow the instructions if the manager pays a cost to them. For different states, the cost of entities is dynamic since the consumption of vehicles is different and the demands of the workers are changing.

\textbf{MDP.} Our setting builds on the standard formulation of the Markov Decision Process (MDP) \cite{sutton1998introduction} where the agent observes a state $s\in \mathcal{S}$ and takes an action $a\in \mathcal{A}$. The transition probability function $P(s'|s,a)$ transits current state $s$ to next state $s'$ after taking action $a$, and the agent receives a reward $r$ according to the reward function $r:\mathcal{A}\times\mathcal{S}\to\mathbb{R}$. The goal of the agent is to learn a policy $\pi(a|s)$ that maximizes the expected cumulative discounted returns $\mathbb{E}_{\pi}\left[\sum_{t=0}^{\infty} \gamma^{t} r\left(s_{t}, a_{t}\right)\right] \text { where } \gamma \in[0,1) \text { is the discount factor}$. \textcolor{black}{In this paper, the state of the manager comprises the positional information of all entities, attributes of carried resources, the positions of tasks, and the total resources required for tasks. Actions are defined as the task allocation scenario. The state transition matrix is based on the current task allocation scenario, where the environment returns the positions and resources of all entities and tasks after each entity's movement. The state of an entity includes its own resources and position, as well as the resource requirements and positions of all tasks. Actions involve moving in any direction, and for worker entities, there is an additional action for bidding. The state transition matrix is determined by the current state and action, returning the entity's resource position and the resource requirements and positions of all tasks for the next time step.}

\textbf{Task.} Let $T = <T_1,T_2,...T_m>$ be the sequence of tasks arriving in an episode, and task $T_i$ is defined as a tuple $<r_i,tr_{i,1},tr_{i,2},...tr_{i,k}>$ where $r_i$ denotes the reward of completing task $T_i$ and $tr_{i,j}$ is the quantity of resource $j$ needed to complete task $T_i$. Ability, experience, or the number of items can all be interpreted as resources. We define $C_i$ as the flag of completion of task $T_i$, i.e., $C_i=1$ if task $T_i$ is completed, otherwise $C_i=0$. 

\textbf{Entities.} 
\textcolor{black}{In this paper, we extend the concept of reinforcement learning agents to entities, which can be considered roughly equivalent. Entities can be item entities (e.g., vehicles) or worker entities. The difference between the two lies in whether the entity can provide bids based on its own attributes. The former can be regarded as fixed-price untrained agents, while the latter employs reinforcement learning algorithms to dynamically adjust bidding strategies.} In the subsequent descriptions, we will use the term "entity" instead of "agent." Item entities have a cost (e.g., fuel consumption and time penalty), while worker entities have wage demands for task assignments. Each entity $w_i$ is defined as $<d_i, wr_{i,1}, wr_{i,2}, ..., wr_{i,k}>$, with $d_i$ representing the cost demand and $wr_{i,j}$ denoting the resource $j$ possessed by entity $w_i$.  For item entities, $d_i$ is a manually given cost function based on distance, time, purchase cost, etc. For Worker entities, they aim to maximize their individual rewards. Workers propose demands $d=<d_1,d_2,...d_n>$ based on their resources $wr = <wr_1,wr_2,...wr_k>$. A high demand reduces the chances of selection, while low demands result in lower task rewards. Worker entities take a policy $\pi_i(d_i|T)$ to determine appropriate reward demands.

\textbf{Manager.} Our algorithm acts as a manager, assigning entities to tasks based on task and entity attributes. The manager observes the task $T$, entity resources $wr$, entity costs $d$, and chooses entities ($a=<a_1,a_2,...a_m>$) to perform tasks. The manager's goal is to maximize its own reward, $\sum_{i=1}^{m}R_{i}C_i$, where $R_{i}=r_i-\sum_{j\in a_i}d_j$ represents the reward for task $T_i$ with action $a_i$. If $R_{i}<0$, the manager abandons task $T_i$ ($C_i=0$) due to negative profit. To maximize returns, the manager ensures selected entities have sufficient resources for task completion and minimizes entity selection costs. A policy $\pi(a|T,w)$ is learned to allocate entities to suitable tasks.
\begin{figure*}[t]
 \begin{center}
  \includegraphics[height=240pt,width=400pt]{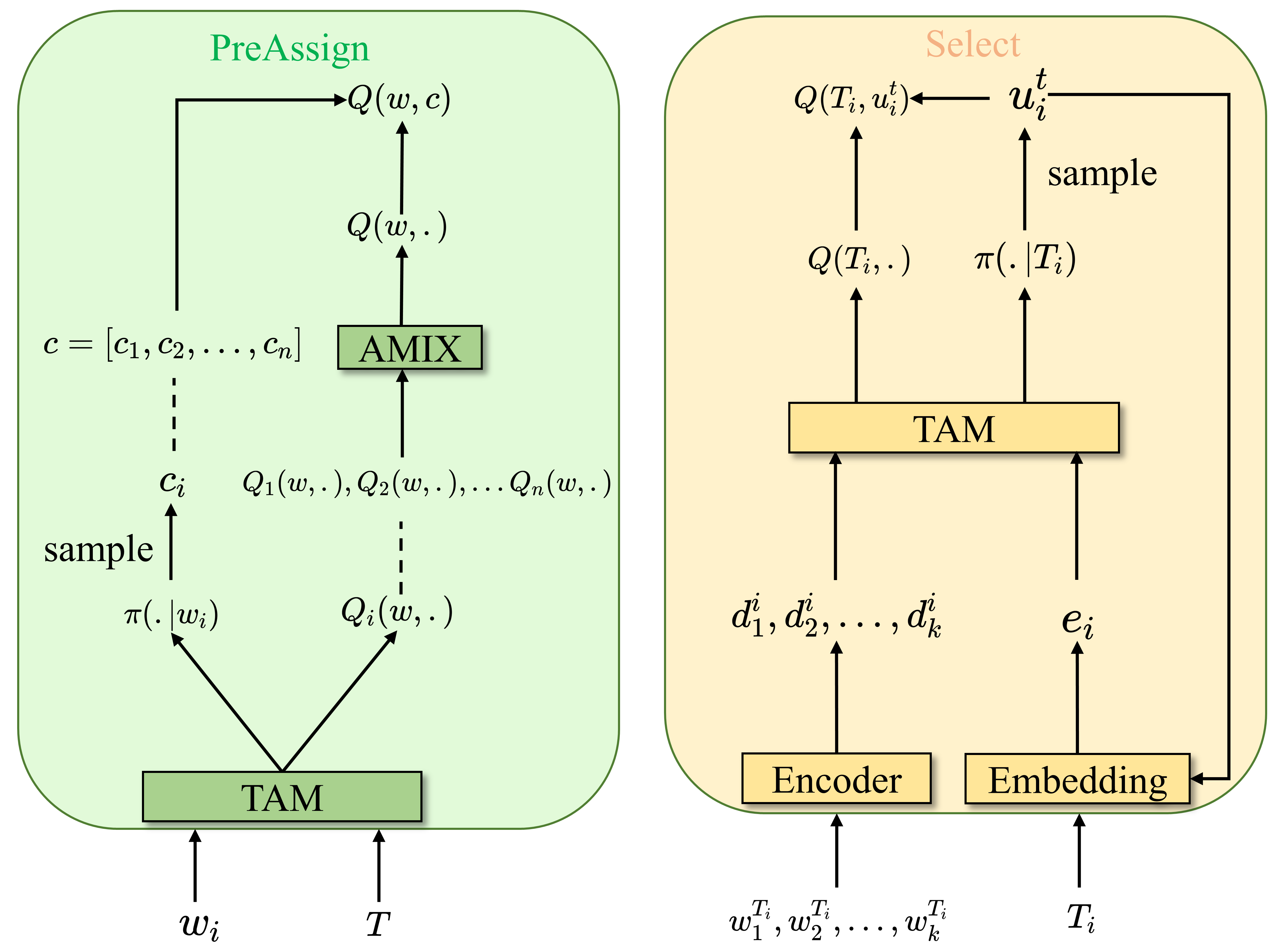}
  \caption{The architecture of the two-stage approach. We employ the PreAssign model to preselect candidate entities for each task and subsequently choose the entities to perform each task.}
  \label{Architecture}
   \end{center}
\end{figure*}

\begin{figure*}[t]
  \centering
  \subfigure[]{
    \begin{minipage}[t]{0.56\linewidth}
    \centering
    \includegraphics[height=200pt,width=260pt]{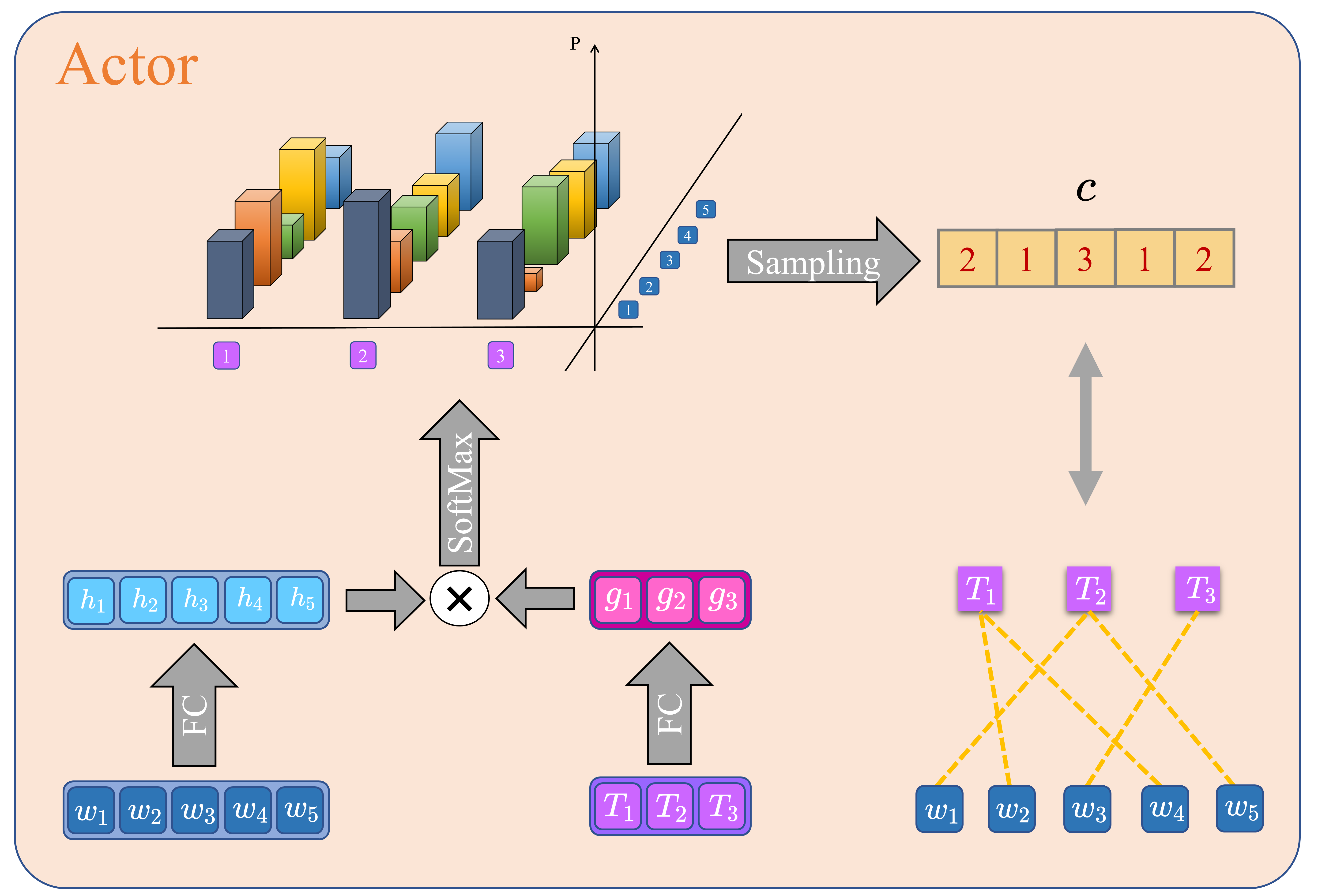}
    \end{minipage}
    }
    \subfigure[]{
    \begin{minipage}[t]{0.4\linewidth}
    \centering
    \includegraphics[height=200pt,width=180pt]{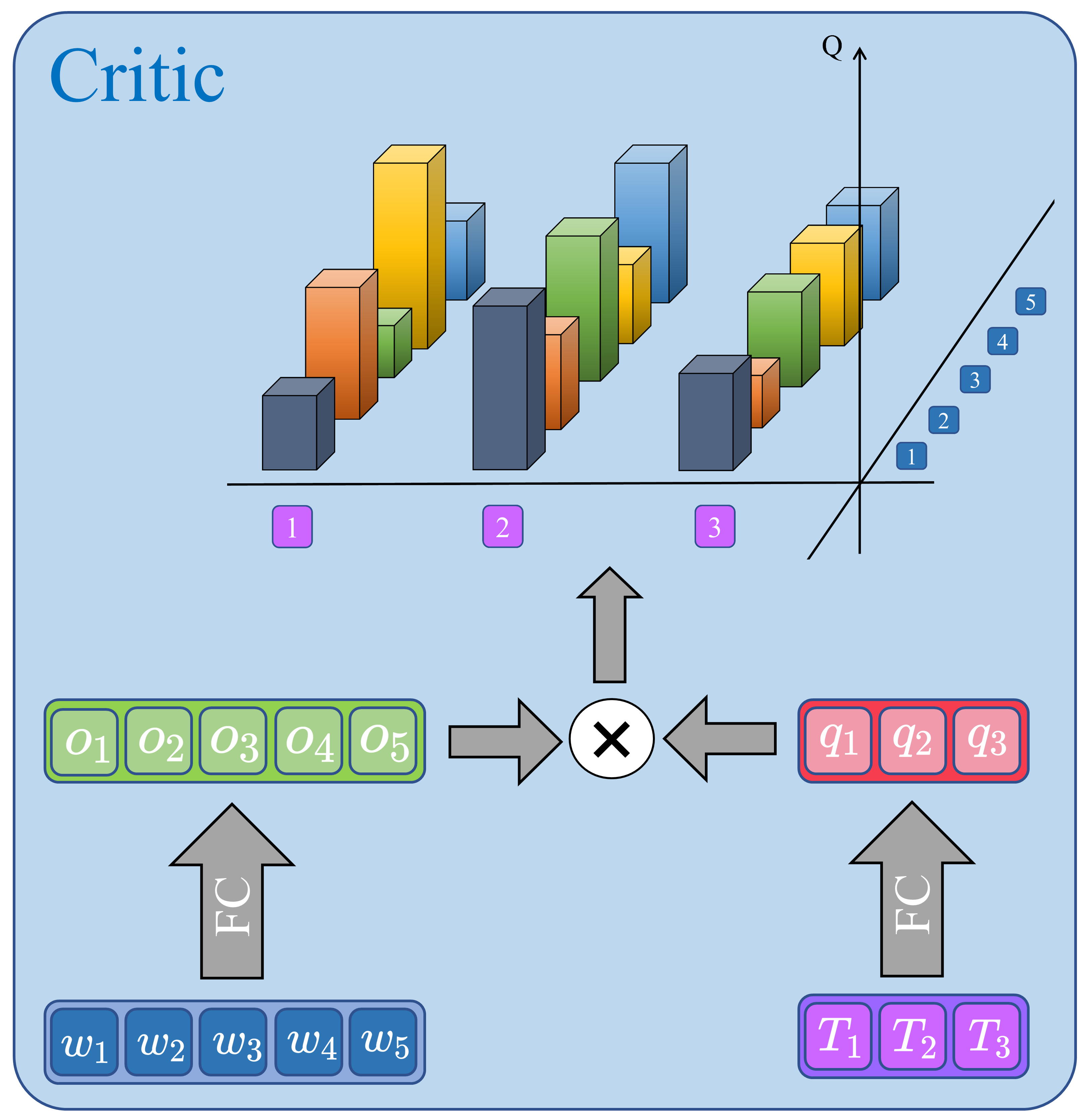}
    \end{minipage}
    }
  \caption{The Two-head Attention Module. The actor and the critic are both computed by the two-head attention module. (a) is the actor's component, and (b) is the critic's component. The process begins with the embedding of entities and tasks. Entities are embedded as queries, while tasks are embedded as keys. The dot product of queries and keys is then computed to derive a score, which is utilized to create a distribution of pre-assigned actions in the first head and estimate the value of each action in the second head. By sampling an action from this distribution, a pre-assigned action is determined, and its value is estimated by the second head.}
  \label{PreAssign}
\end{figure*}

\section{Methodology}

\begin{figure*}[t]
    \centering
    \includegraphics[height=200pt,width=420pt]{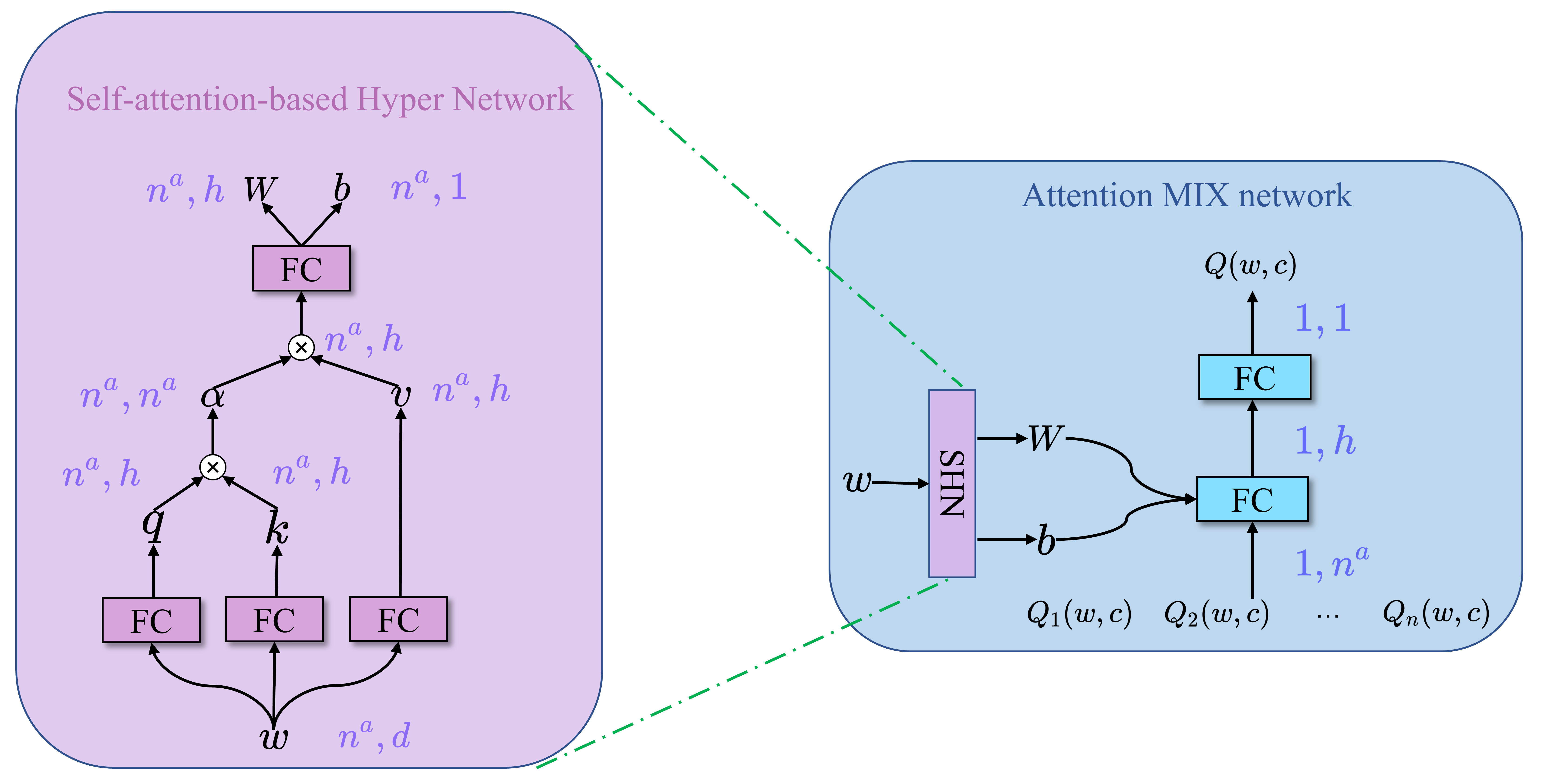}
    \caption{The architecture of AMIX. The Self-attention-based Hyper Network (SHN) can generate variable sizes of parameters by using attention mechanisms. The input of SHN is the attributes of entities, and the outputs are the parameters of the first layer in AMIX. \textcolor{black}{When there are $n^a$ input entities, SHN will output weight parameters of dimension $n^a \times h$ and bias parameters of dimension $n^a \times 1$. Then, AMIX will take the $1 \times n^a$ dimension Q-values as input, multiply them by the weights of SHN, and add the bias to obtain the output of the first layer network, which has a dimension of $1 \times h$. Regardless of the number of entities, the final output dimension remains the same. This allows for flexible allocation planning with varying numbers of entities. The second layer is a fully connected layer and finally outputs the estimated total value.}}
  \label{AMIX}
\end{figure*}

The manager faces challenges in allocating entities to tasks. One challenge is the uncertainty in the number of tasks, making it difficult to allocate entities simultaneously. To address this, we propose setting the action dimension as $n\times M$, \textcolor{black}{where $M$ denotes the maximum number of tasks. However, obtaining the precise value of $M$ in advance is challenging, and in most cases, the total number of tasks is smaller than the maximum number of tasks, which can result in wasted space.} Another challenge is determining the task allocation order. Fixed order may result in suboptimal solutions, while simultaneous allocation makes it difficult to decide which task each entity should be assigned to. Additionally, the manager needs to consider the generalization problem, where entity costs and resources can vary. \textcolor{black}{Most RL algorithms are unable to handle the issue of variable numbers of agents. The majority of RL algorithms typically assume a fixed environment, meaning a fixed number of agents. This assumption makes these algorithms ill-suited for handling problems with variable numbers of agents. For instance, when the number of agents dynamically changes, traditional reinforcement learning algorithms may struggle to adapt effectively because they often require a predefined number of agents and state spaces. Therefore, when the number of agents fluctuates, these algorithms may encounter difficulties, leading to degraded performance or failure to converge.}

To address these challenges, we propose a two-stage approach for entity selection. \textcolor{black}{
We first propose a pre-allocation method where each entity is pre-assigned as a candidate for each task based on its attributes and the attributes of the tasks. This method avoids the drawbacks of sequential and random selection by enabling sequential allocation based on the fitness between tasks and entities. We have also demonstrated that this pre-allocation method outperforms sequential allocation methods in finding optimal solutions. Secondly, our proposed attention-based hyper network structure allows for the allocation of different network parameters to different numbers of entities, enabling the neural network to accommodate dynamic entity counts and addressing the generalization issue when the number of entities changes.} It overcomes local convergence and entity overlap issues and handles generalization problems. We introduce worker entities as selfish agents to verify the allocation of entities with varying attributes. The overview of our method is depicted in Figure \ref{overview}, while the architecture is illustrated in Figure \ref{Architecture}.

\subsection {Pre-assign Module}

\begin{figure*}[b]
    \centering
    \includegraphics[height=240pt,width=440pt]{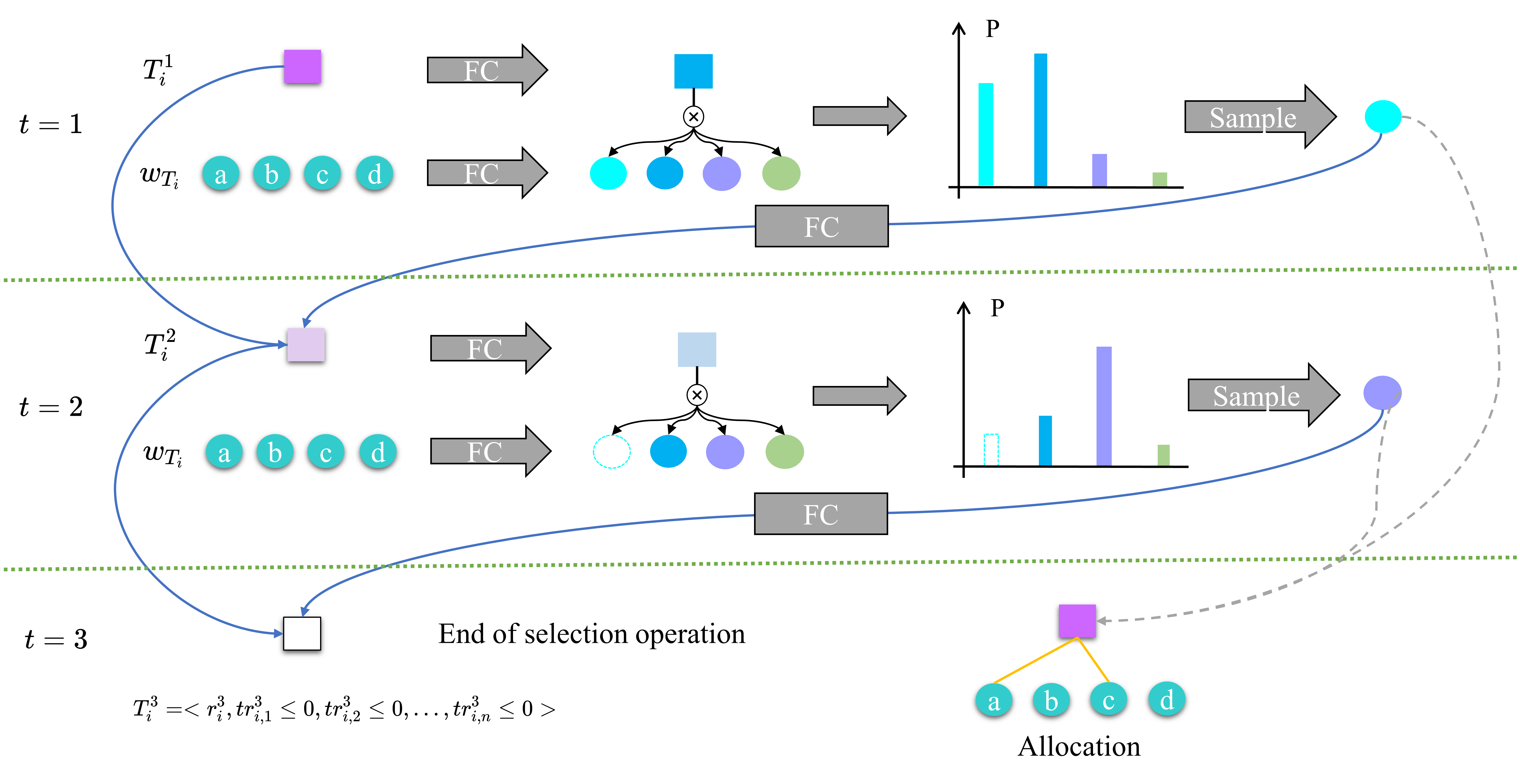}
    \caption{The example of how select module chooses entities. The entities and tasks are embedded, and then the attention mechanism is used to compute the similarities. Similar to the method that modifies the hidden state of the sequence to indicate the word has been chosen, the entity is selected by sampling, and then it will affect the task in the following selection step to denote this entity has been chosen.}
  \label{Select}
\end{figure*}

The pre-assign module initially allocates entities to suitable tasks. In discrete action space environments, each discrete number typically corresponds to an entity action. However, traditional RL algorithms focus on learning the likelihood of number occurrences in the current state, which limits their ability to effectively utilize information in task assignment problems. To leverage the individual attributes of each entity, we propose a Two-head Attention Module (TAM) based on attention mechanisms (\cite{vaswani2017attention}). This module guides the manager in making pre-assigned allocations based on the compatibility between tasks and entities. It provides policy and assignment value outputs, and its network structure accommodates a variable number of input entities. \textcolor{black}{We adopt Soft Actor-Critic method (SAC, \cite{haarnoja2018soft}) as the base RL algorithm for modeling the manager. Since the action space for task allocation is discrete, we adopted the discrete version of the SAC algorithm (\cite{christodoulou2019soft}).}

TAM consists of two heads: the actor head and the critic head. The actor head is responsible for generating the policy for the pre-assign actions, while the critic head calculates the value of these actions. The actor head consists of an entity embedding function $f^{h}: \mathbb{R}^{S} \rightarrow \mathbb{R}^{d}$, a task embedding function $f^{g}: \mathbb{R}^{S} \rightarrow \mathbb{R}^{d}$.  Here, $S$ represents the dimension of the task and entity attributes. The entity embedding can be expressed as $\boldsymbol{h}_i=f^{h}(w_i)$ and the task embedding can also be denoted as $\boldsymbol{g}_i=f^{g}(T_i)$. The probability of each entity pre-assigning to each task is computed by an attention module, i.e., $\pi(T_j|w_i)=\text{SoftMax}(\boldsymbol{h}_i^T\boldsymbol{g}_j/\sqrt{d})$. This method can solve the problem of a variable number of tasks with zero-shot training because when an unseen task $T_u$ occurs, we can compute the embedding $\boldsymbol{g}_u=f^g(T_u)$ and then the probability of entity $w_i$ assigning task $T_u$ is $\pi(T_u|w_i)=(\boldsymbol{h}_i^T\boldsymbol{g}_u/\sqrt{d})$. The pre-assign allocation of entity $w_i$ is $c_i$, which is sampled from the $\pi(T_j|w_i)$. The total pre-assign allocation is $c=(c_1,c_2,...c_n)$ where $c_i=k$ denotes that $w_i$ is pre-assigned to task $T_k$. 

Due to the discrete nature of pre-assigned actions, a table can be created to record each action's value. However, the table size ($2^n$) becomes impractical when $n$ is large, making it challenging to accurately approximate values with limited datasets. To address this, we adopt an approach inspired by \textcolor{black}{the Q MIXing network (QMIX) \cite{rashid2020monotonic}, which employs a mixing network to compute the total Q value for multi-agent systems. In our approach, we calculate the value selected for each entity and consider the overall assigned value as a nonlinear aggregation of the individual entity values.}
\textcolor{black}{
In QMIX, each agent has a Q network that calculates the value of taking a particular action based on the current observations, resulting in a total of as many Q values as there are agents. Additionally, QMIX employs a hyper network that generates non-negative parameters for the neural network, combining each Q value to compute the overall Q value.} For algorithm fine-tuning, we modify the attention mechanism by incorporating a critic head, similar in architecture to the actor head, to output action values. The critic component includes entity value embedding ($f^{o}$) and task value embedding ($f^{q}$) functions denoted as $\boldsymbol{o}_i=f^{o}(w_i)$ and $\boldsymbol{q}_i=f^{q}(T_i)$, respectively. The pre-assign value of $w_j$ to $T_i$ is computed using the dot product, $Q(w_i,.)=\boldsymbol{o}_i^T\boldsymbol{q}$. The architecture of TAM is depicted in Figure \ref{PreAssign}.

When a pre-assigned action of an \textcolor{black}{entity} is sampled, we obtain a value from the critic's component. In the case of having $n$ \textcolor{black}{entities}, we obtain $n$ values corresponding to the pre-assigned actions. However, since we interact with the environment throughout the entire pre-assign action, we only receive a single total reward from the environment, which does not align with the output of our critic. To address this disparity, we introduce the Attention MIXing (AMIX) module. The AMIX module utilizes a self-attention-based hypernetwork to generate parameters for its layers. \textcolor{black}{Unlike QMIX, which provides network parameters for a fixed number of agents, AMIX's hyperparameter network, called the Self-attention-based Hyper Network (SHN), utilizes an attention model. This allows it to dynamically adjust the number of parameters based on the input values, effectively accommodating the variable number of entities in the problem. For a visual representation of this concept, please refer to Figure \ref{AMIX}.} 
\textcolor{black}{Using the aforementioned critic structure, we can calculate the value estimates for task allocation. According to the SAC algorithm, the actor loss can be expressed as:}
\textcolor{black}{
\begin{align*}
 \mathcal{L}_\text{actor}(\theta) = \alpha \log \pi_\theta(\mathbf{c}|\mathbf{w})-Q_\varphi(\mathbf{w},\mathbf{c})
\end{align*}
}
\textcolor{black}{After the allocation is complete, the critic is updated based on the reward values obtained from the interaction between the agent and the environment. The loss can be expressed as:}
\textcolor{black}{
\begin{align*}
\mathcal{L}_\text{critic}(\varphi)=\frac{1}{2}\left\|r(\mathbf{w}, \mathbf{c})+\gamma \arg \max _{\mathbf{a}} Q_\varphi\left(\mathbf{w}^{\prime}, \mathbf{a}\right)-Q_\varphi(\mathbf{w}, \mathbf{a})\right\|^{2}
\end{align*}
}
\textcolor{black}{Through the aforementioned loss, we can continuously update the critic network to estimate the reward values for the current allocation executed by all entities. This allows the actor network to output optimal task allocations.}

\subsection {Select Module}

When the \textcolor{black}{entities} are already pre-assigned to tasks by using the pre-assign module, there are many \textcolor{black}{entities} for each task to select. The manager needs to select the proper number of entities with different resources to complete the task. In order to have better generalization, the selection module must be able to assign \textcolor{black}{entities} and tasks that have not been seen before. To solve these two problems, referring to the point-net (\cite{vinyals2015pointer}) network framework, we establish a seq2seq-like network structure as our select module. There are several reasons why the seq2seq-like structure can solve our problems. 1) The seq2seq structure is widely used for machine translation and is able to generate a variable number of outputs according to the input information. The problem of a variable number of \textcolor{black}{entities} can be solved. 2) In the seq2seq structure, the information of the previous output is taken into account in the context, which helps to achieve better output results in the future. This structure can also help the manager choose \textcolor{black}{entities} better by considering the context of previous selections. 3) The point-net attention mechanism can solve the out-of-vocabulary problem when the input and output are the same. The input and output are both \textcolor{black}{entities} in the allocation problem.

The architecture of the select module is shown in Figure \ref{Select}. The input of the encoder is the \textcolor{black}{entities} $w_{T_i}$ pre-assigned to the task $T_i$, and the input of the decoder is $T_i$. Note that there is no sequential relationship between input \textcolor{black}{entities} like text, so unlike traditional seq2seq structure, we did not use the commonly used RNN and attention mechanisms to encode the input. We simply use fully connected layer as the encoder, and the output for \textcolor{black}{entity} $w_{T_i}$ is $d_{i}$. The embedding of the task $T_i$ is $e_i$, and the output of the $t$-th \textcolor{black}{entity} $u_i^t$ is selected by sampling from a distribution which is $\pi(.|e_i^t)=\text{SoftMax}(v^{T} \tanh (W_{1} e_i^t+W_{2} d^i))$. The previous output will impact the context, so we change the task by $T_i^{t+1} = T_i^t + f^a(u_i^t)$. When the task resources needed for $T_i^t$ are all zero, the process of selection will end.

\subsection {Demand Module}
The worker entities have their own minds, which means they will adjust their demands in different situations. We model workers as entities that can propose their own demands based on the attributes of the task. They can update their own minds by using RL algorithms. Since each entity only considers maximizing their own reward and does not consider the global reward, and multi-agent RL algorithms focus on the optimal overall reward, which is not consistent with our scenario, we use singe-agent RL algorithm to train each entity. The demand of each agent is a continuous number, so we use the DDPG (\cite{lillicrap2015continuous}) algorithm to train these entities. \textcolor{black}{The observation in DDPG is defined as the current attributes and current position of the entity, while the action is defined as the bid value for the demand. When the current entity is selected, the reward obtained is the bid value. When not selected, the reward value is 0. Therefore, worker entities need to provide reasonable bids based on their current attribute values to be selected and obtain high returns.}

\section{Experiment}

In this section, we try to answer these questions: (1) Is the pre-assign method more effective than the sequential task allocation method in avoiding finding local optimal solutions? (2) Can the algorithm solve the allocation problem when the cost of entities is dynamically changed based on the current task? (3) Can the algorithm provide a reasonable allocation for such a variable number of entities as their number increases and decreases? (4) Does our proposed method outperform heuristic algorithms?
\textcolor{black}{For all experimental environments, our two-stage method uses the same hyperparameters, with no parameters specifically set for each environment. For all network architectures, we use the ReLU function as the activation function and set the number of hidden units to 64, with each network consisting of a 3-layer MLP structure. More details on the parameter configurations of our method can be found in Table \ref{param}.}

\subsection{Retain the Almighty}
% \begin{figure*}[t]
%     \centering
%     \subfigure[]{
%     \begin{minipage}[t]{0.5\linewidth}
%     \centering
%     \includegraphics[height=70pt,width=160pt]{imgs/o.png}
%     \end{minipage}
%     }
%     \subfigure[]{
%     \begin{minipage}[t]{0.4\linewidth}
%     \centering
%     \includegraphics[height=70pt,width=120pt]{imgs/match.png}
%     \end{minipage}
%     }
%     \caption{The Retain the Almighty Environment. Left: each task has an agent set and the task is done if one entity in this set is selected. The last task can only be completed if the Almighty is selected. The right picture shows the return of this environment by using the sequential method, the random order method, and the pre-assign method.}
%   \label{Almighty}
% \end{figure*}

This small experiment will answer the question (1) and demonstrate why using sequential selection methods can fall into local optimization and prove that using the pre-assign method is more effective.
\begin{wrapfigure}{0.4\textwidth}{0.6\textwidth}
\centering
\includegraphics[width=0.6\textwidth]{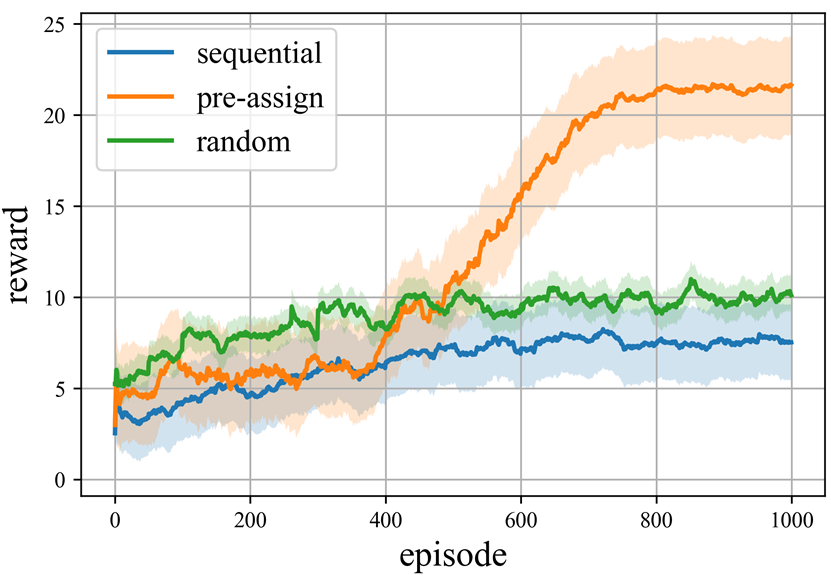}
\caption{The Retain the Almighty Environment. The picture shows the return of this environment by using the sequential method, the random order method, and the pre-assign method.}
\label{Almighty}
\end{wrapfigure}

In the Retain the Almighty environment, there are $N$ tasks and $n$ entities in this environment, and each task $T_i$ in this environment corresponds to a subset of the best entities $b_i$, and the task is completed if and only if a best entity $w\in b_i$ participates in this task. The last task $T_N$ is a difficult but rewarding task. The best entity of $b_N$ is $w_o$, which is an almighty agent that is able to finish every task, and the cost for it is the same as others. The best allocation is to arrange $w\in b_i$ to do $T_i$ and the task cannot choose the entities if the previous task has chosen them, which means the algorithm must retain the almighty to the last task to find a good solution. The result is shown in Figure \ref{Almighty}.

It can be seen that the \textcolor{black}{entity} learned using the sequential method is getting trapped in the local optimum. The perfect \textcolor{black}{entity} $w_o$ is selected by the previous task and fails to complete the last task, resulting in a low total reward. When the order of the most difficult tasks is randomly assigned to the top, we can assign the Almighty to this task and obtain high profits; when the most difficult task is randomly assigned later, the rewards will sharply decrease. According to the training curve, the \textcolor{black}{entity} has not learned how to assign entities to appropriate tasks, and the final return of this \textcolor{black}{entity} is related to the random order of tasks instead of how well it learns to allocate \textcolor{black}{entities} to tasks. However, by using the pre-assign method, the manager is able to find the best solution because the task can be completed once the best \textcolor{black}{entity} is pre-assigned to the last task. The manager is able to allocate $w_o$ to $T_N$ to get a large reward and then find the optimal solution.

In fact, in this environment, we can suppose that without prior knowledge, all entities have the same probability of being selected when the algorithm does not start training. The probability upper bound of $w_o\in T_N$ by using the sequential method is $(\frac{n-1}{n+N-2})^{N-1}$. The probability upper bound decreases exponentially with the number of tasks. However, the probability of $w_o\in T_N$ by using the pre-assign method is $\frac{1}{N}$. The rate of probability decline is far less than that of the sequential method. See Appendix \ref{APPENDIXA} for complete proof.

\subsection{Electric Power Transportation}
In this experiment, we will answer question (2). There are many towers in the environment that are connected by wires. Different towers have different wire materials, which means different transmission costs between towers. At each time step, there may be several towers in the city corresponding to the power peak, so other towers are needed for power transmission. As a manager, we need to select the proper electric towers to carry out power transmission for the towers with peak power consumption so as to complete the task of ensuring the smooth use of electricity in the city. The towers awaiting transmission represent tasks, characterized by their locations and additional power requirements. Other towers serve as entities, characterized by their additional available power and locations. We employ our proposed two-stage method to initially allocate tasks to each tower awaiting transmission and then make selections based on the specific additional power and attributes of each tower. Once selected to participate in transportation, the cost is calculated as the distance between the two stations multiplied by the predetermined cost of transmission per meter of wire between the two towers. The transmission cost per meter of wire is determined by the material between the two towers, which is predefined in the environment. \textcolor{black}{For each target tower, if sufficient power transmission is successfully obtained, the reward is the task completion reward minus the power transmission cost. Therefore, while ensuring sufficient power, reducing transmission costs is also considered. In this environment, there are a total of 20 electric towers. Each interaction with the environment causes changes in the power values required by the towers. When the power exceeds their individual limits, support from other towers is needed.}

\begin{figure*}[t]
    \centering
    \subfigure[]{
    \begin{minipage}[t]{0.31\linewidth}
    \centering
    \includegraphics[height=90pt,width=150pt]{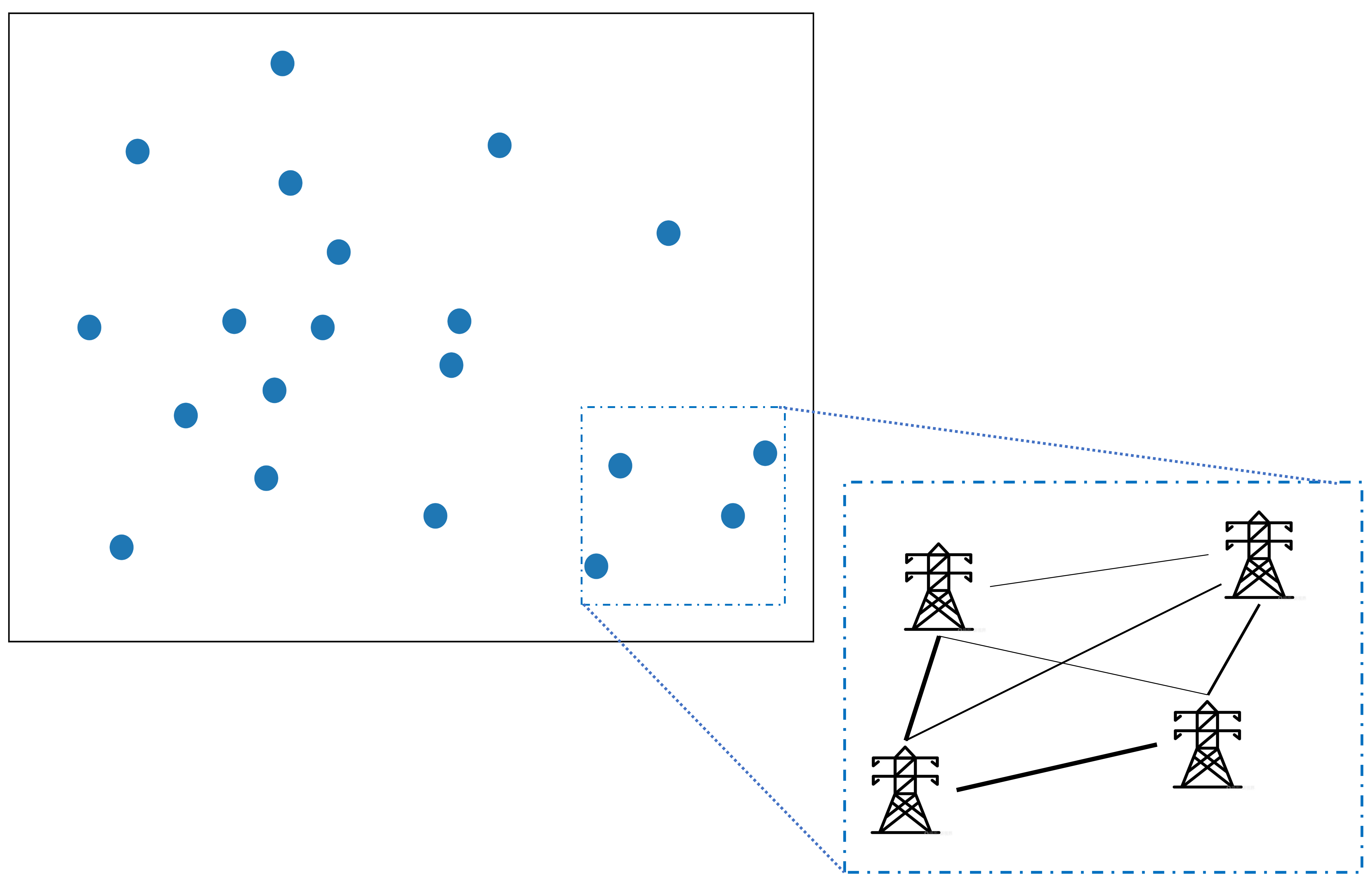}
    \end{minipage}
    }
    \subfigure[]{
    \begin{minipage}[t]{0.31\linewidth}
    \centering
    \includegraphics[height=90pt,width=130pt]{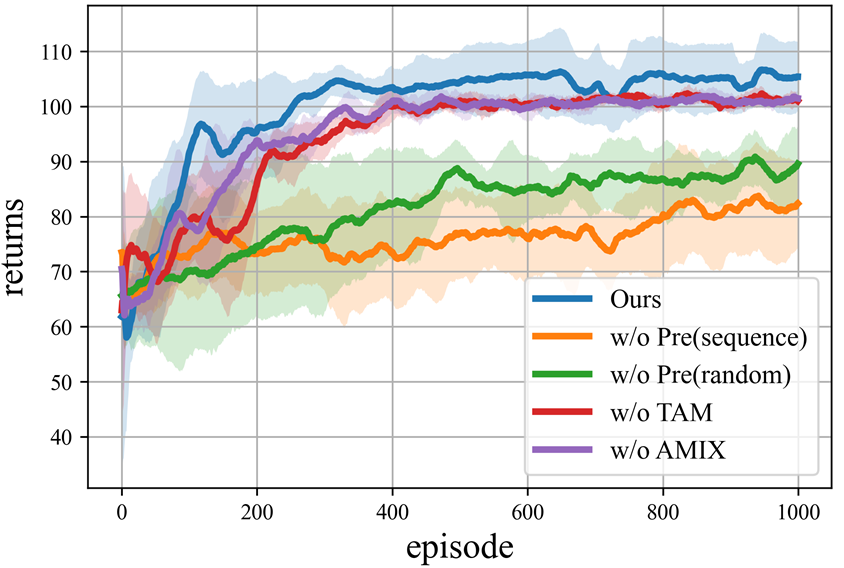}
    \end{minipage}
    }
    \subfigure[]{
    \begin{minipage}[t]{0.31\linewidth}
    \centering
    \includegraphics[height=90pt,width=130pt]{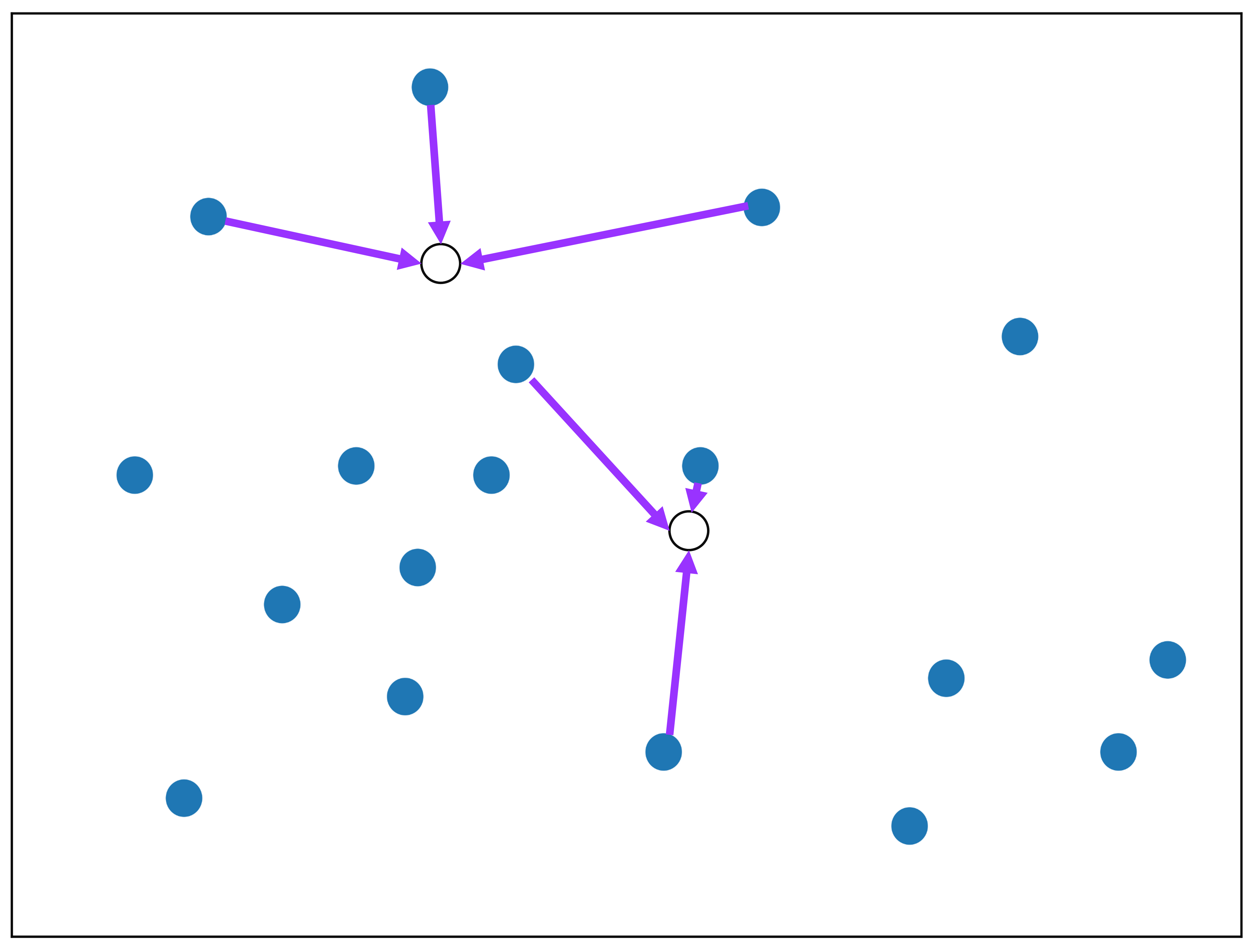}
    \end{minipage}
    }
    
    \subfigure[]{
    \begin{minipage}[t]{0.22\linewidth}
    \centering
    \includegraphics[height=90pt,width=100pt]{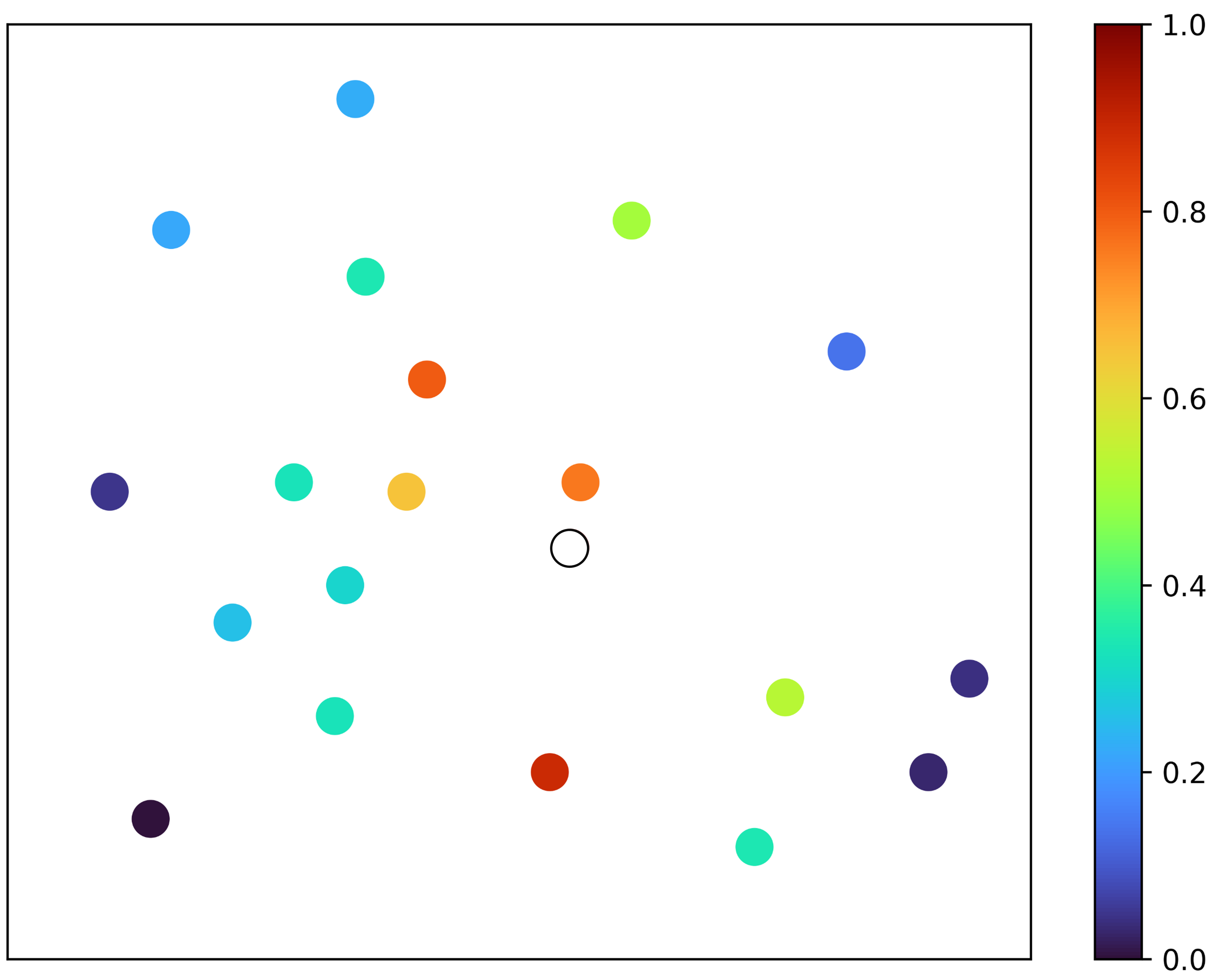}
    \end{minipage}
    }
    \subfigure[]{
    \begin{minipage}[t]{0.22\linewidth}
    \centering
    \includegraphics[height=90pt,width=100pt]{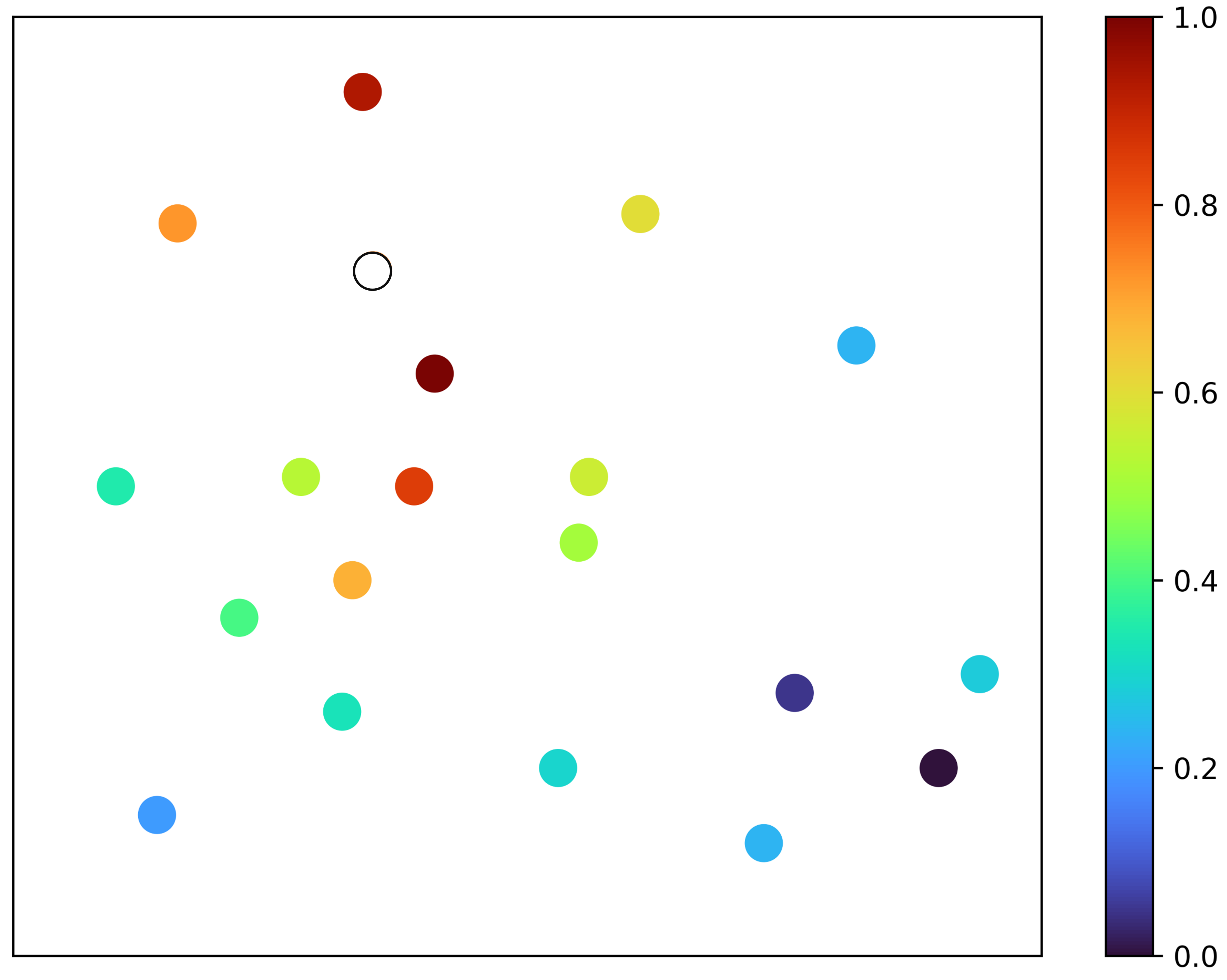}
    \end{minipage}
    }
    \subfigure[]{
    \begin{minipage}[t]{0.22\linewidth}
    \centering
    \includegraphics[height=90pt,width=100pt]{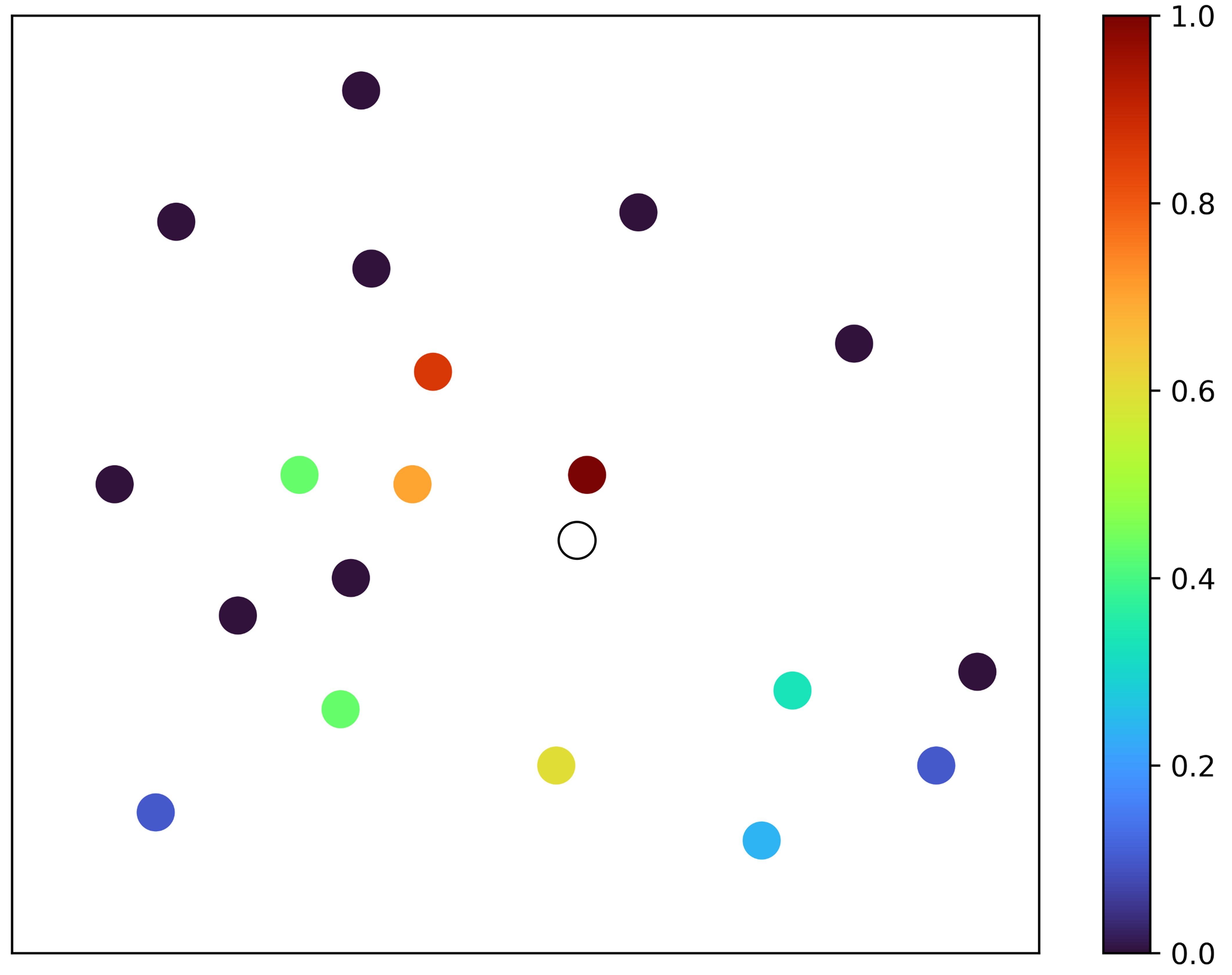}
    \end{minipage}
    }
    \subfigure[]{
    \begin{minipage}[t]{0.22\linewidth}
    \centering
    \includegraphics[height=90pt,width=100pt]{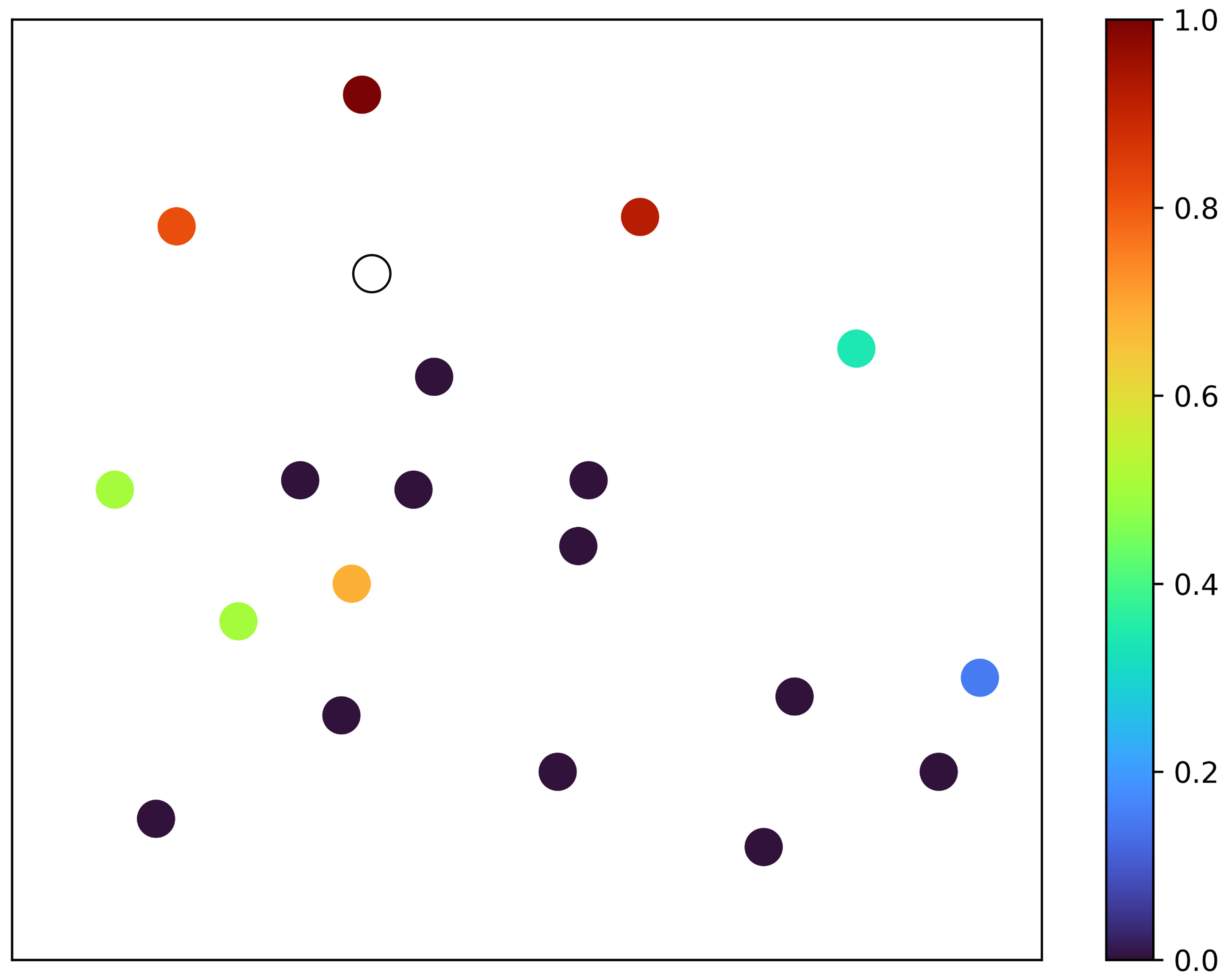}
    \end{minipage}
    }
 \caption{The experiment with Electric Power Transportation. (a) shows the display of the environment. The blue point denotes an electric tower, and each tower is connected by wires. (b) shows the training curve with different methods. \imgg{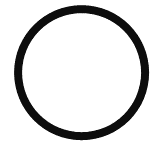} is the task tower, and \imgg{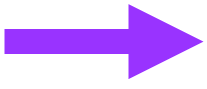} shows the direction of power transmission. The allocation process example of our method is shown in (c)$\sim$(g). (d) and (e) show the value of the pre-assign action, and (f) and (g) show the value of selecting each \textcolor{black}{entity}. The value is normalized to the range of 0 and 1.}
  \label{elec}
\end{figure*}

We use our two-stage approach to train the manager. To test the effect of our method, we denote $w/o \;Pre$ as the architecture without the pre-assign module, $w/o \;TAM$ as the architecture without the attention module, which means the actor and critic are linear layers. The $w/o \;AMIX$ is denoted as the normal global critic to calculate the expected return of this pre-assign action. When we use the AMIX structure, we calculate the individual value assigned to each task for each entity, and then use AMIX to calculate the overall value value; When the AMIX structure is not applicable, we will use an overall critical network to input the attributes of all entities and tasks and directly estimate the overall value value. The training curve and an example of our allocation method are shown in Figure \ref{elec}. \textcolor{black}{From the figure, it can be seen that our proposed method significantly improves the effectiveness of power allocation and increases the reward value. Additionally, the effectiveness of each proposed module is validated through the curve chart. These subplots show two towers at peak electricity usage, requiring other tower entities to transmit power. Subplot (d) illustrates the value assigned to other tower entities when pre-assigned to a central peak-usage tower. Subplot (e) shows the value for a tower located in the upper left, also at peak usage, receiving pre-assign support from other towers. After training, the pre-assign schemes with higher values are distributed around these two target towers, while towers farther away have lower values. This is due to increased transmission costs with distance, making it inefficient to pre-assign distant towers for support. Our method captures this information, enabling better pre-assignment to towers with lower transmission losses. Subplots (f) and (g) reflect the visualization results of the select stage. After pre-assignment, most tower entities are pre-assigned to the nearest peak-usage towers, and each target tower only needs to select from nearby towers. The value in the selection stage shows that target towers still prioritize nearby entities with lower transmission costs, proving that our algorithm can select suitable entities based on pre-assign results to complete the task.}

\begin{figure*}[t]
    \centering
    \subfigure[]{
    \begin{minipage}[t]{0.26\linewidth}
    \centering
    \includegraphics[height=120pt,width=120pt]{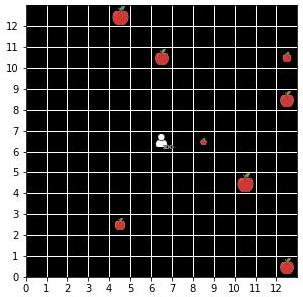}
    \end{minipage}
    }
    \subfigure[]{
    \begin{minipage}[t]{0.26\linewidth}
    \centering
    \includegraphics[height=120pt,width=120pt]{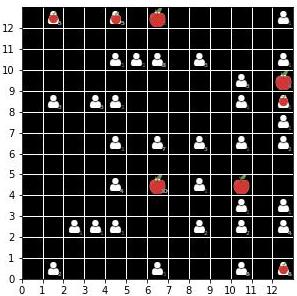}
    \end{minipage}
    }
    \subfigure[]{
    \begin{minipage}[t]{0.4\linewidth}
    \centering
    \includegraphics[height=120pt,width=180pt]{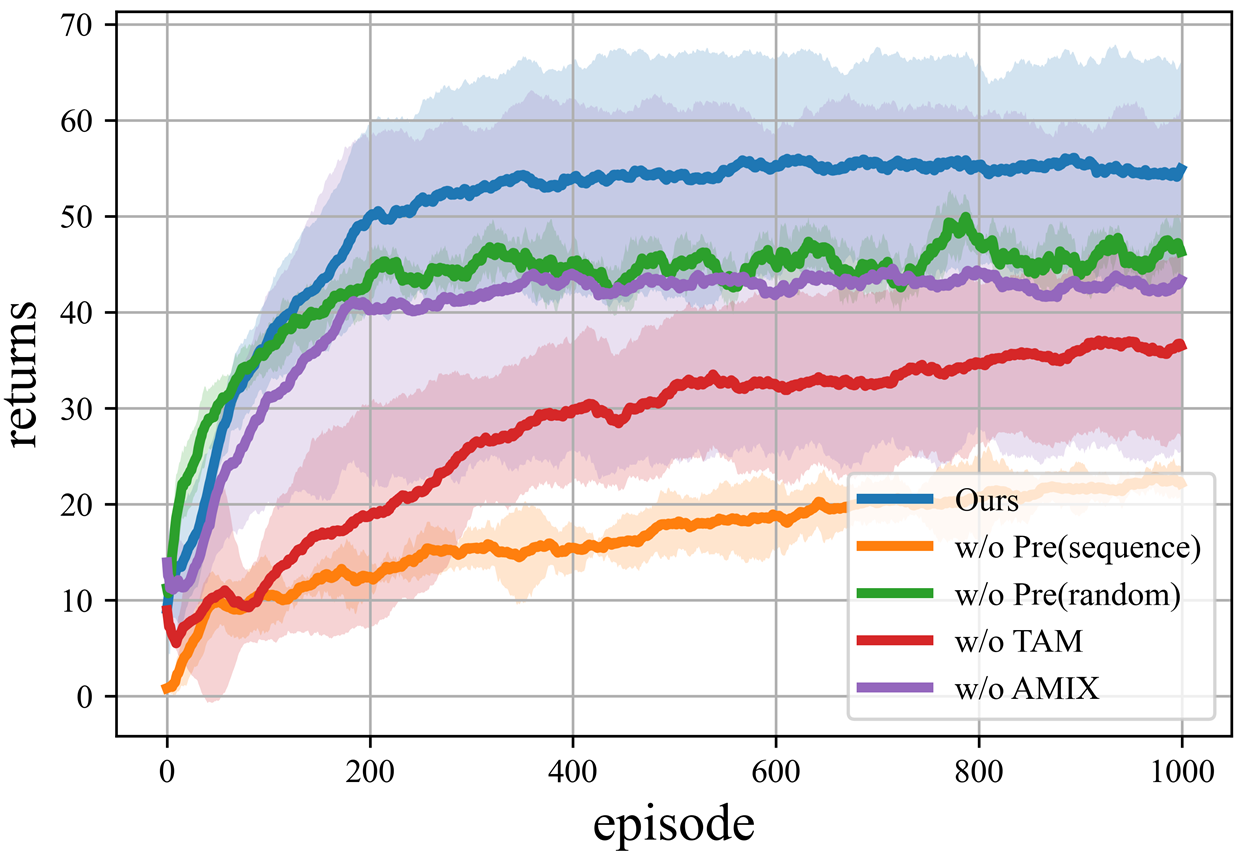}
    \end{minipage}
    }
    
    \subfigure[]{
    \begin{minipage}[t]{0.31\linewidth}
    \centering
    \includegraphics[height=110pt,width=150pt]{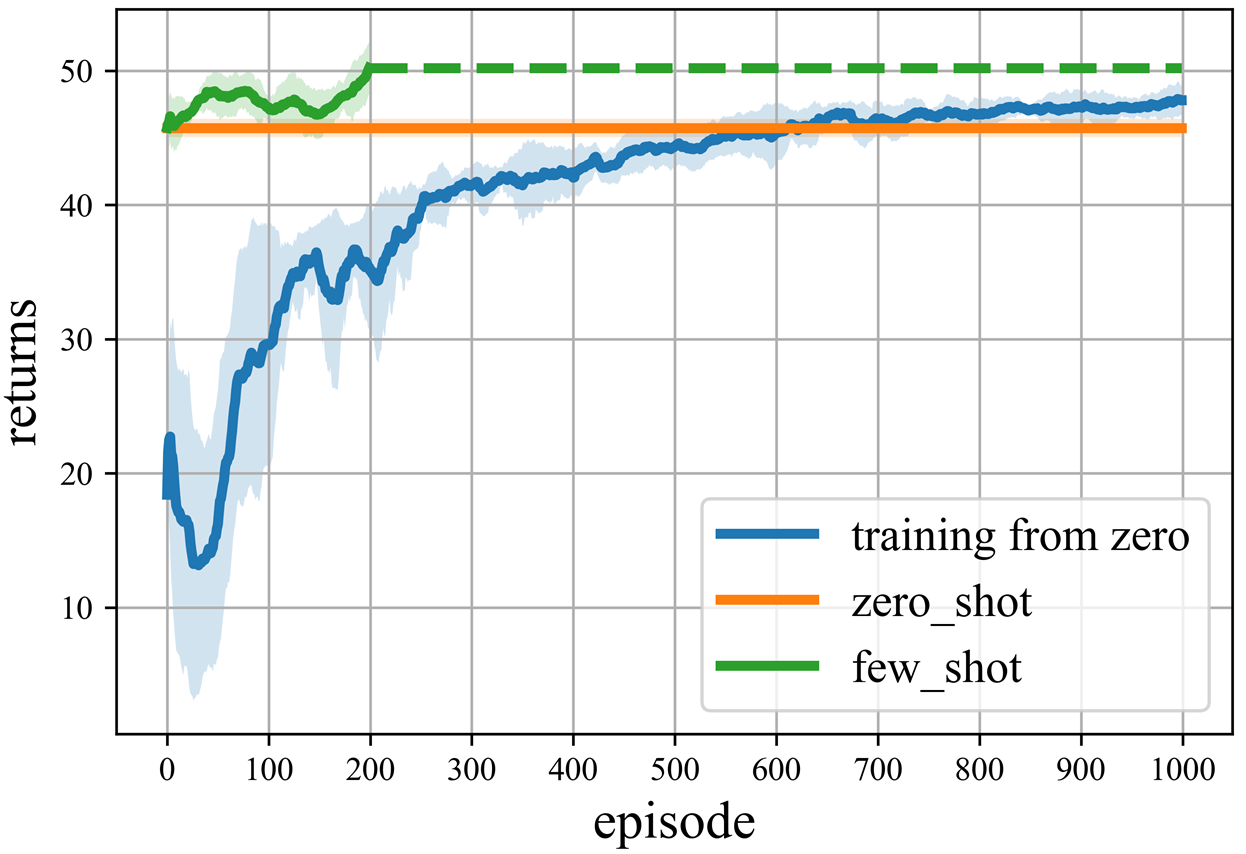}
    \end{minipage}
    }
    \subfigure[]{
    \begin{minipage}[t]{0.31\linewidth}
    \centering
    \includegraphics[height=110pt,width=150pt]{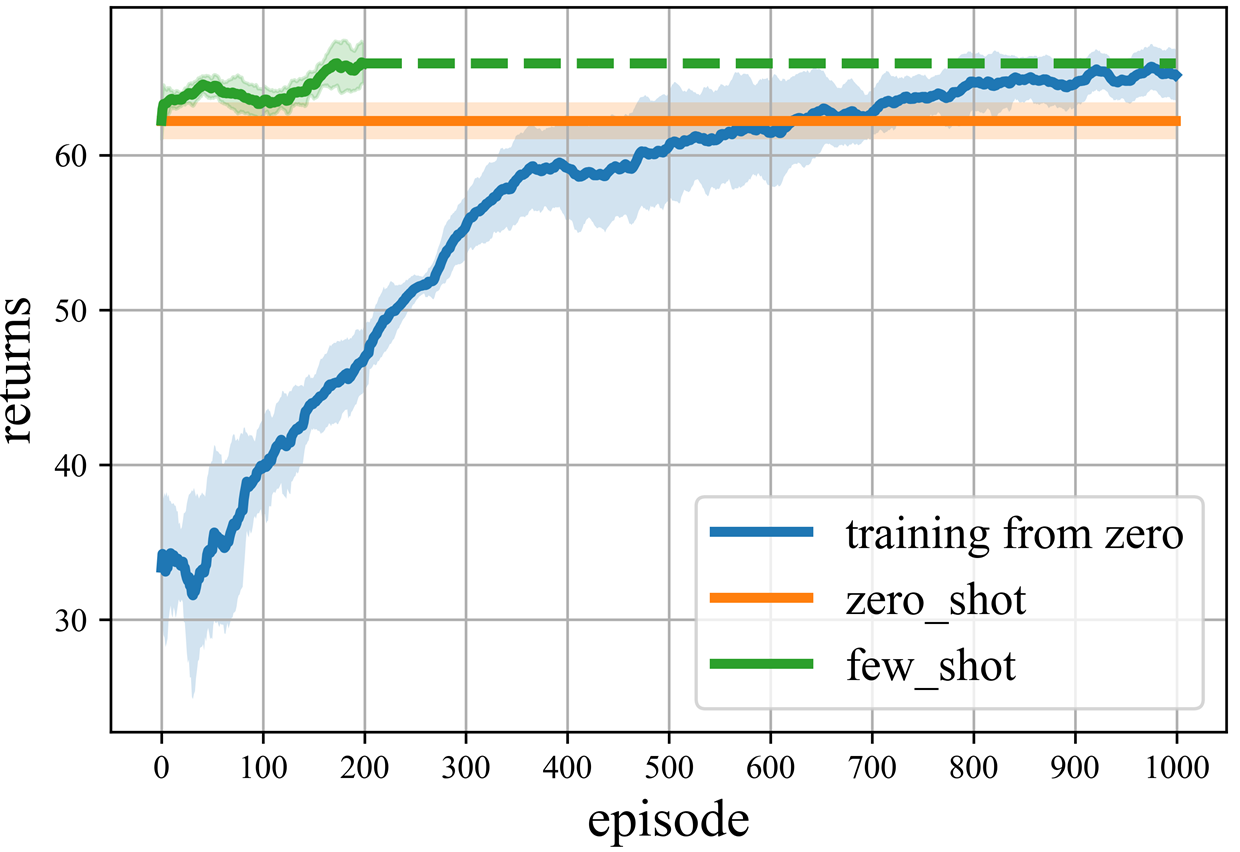}
    \end{minipage}
    }
    \subfigure[]{
    \begin{minipage}[t]{0.31\linewidth}
    \centering
    \includegraphics[height=110pt,width=150pt]{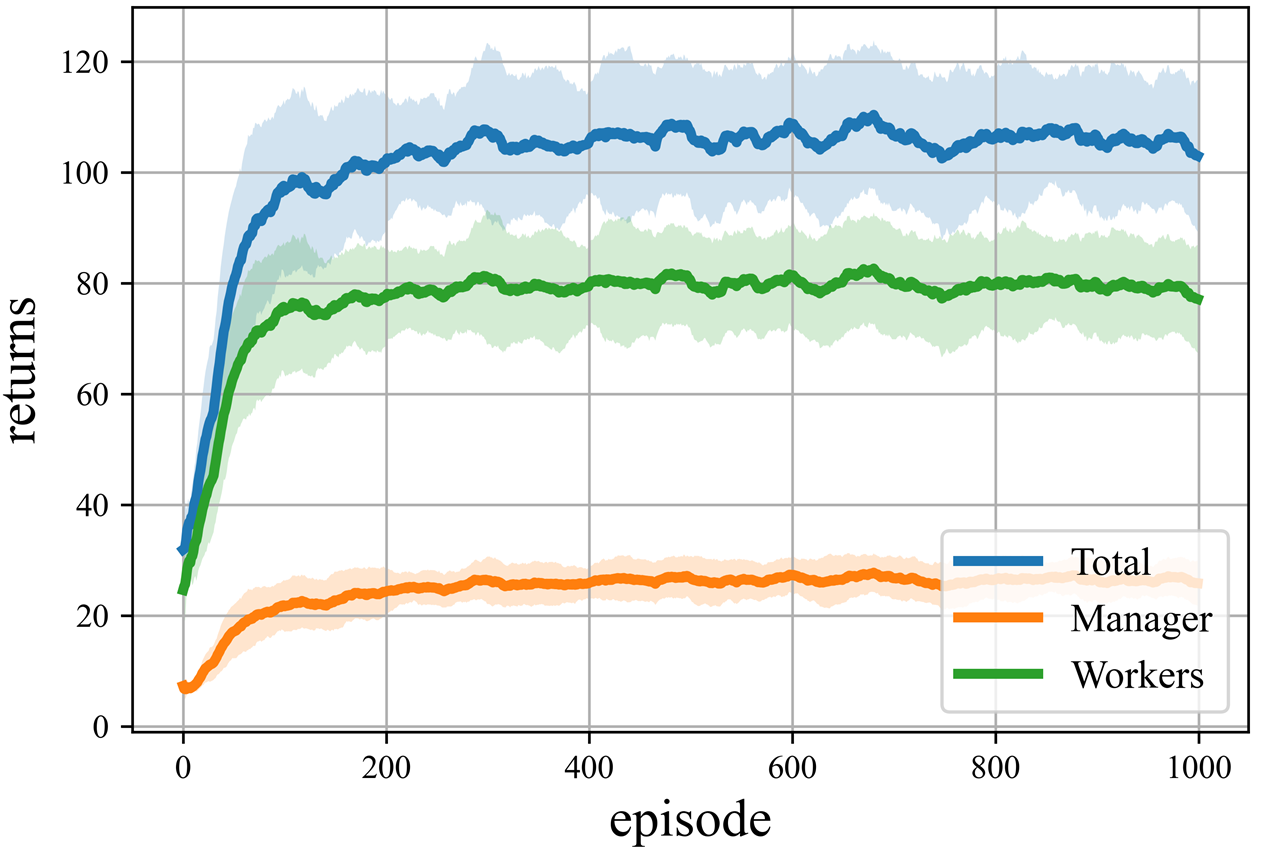}
    \end{minipage}
    }

    \caption{The experiment of RBF. (a) shows the RBF environment. At first, all entities are in the middle of the map. With the emergence of Apple tasks, managers assign entities to complete these tasks, and (b) shows the positions of all entities at a certain moment and the current task positions. The number in the bottom right corner of the entities indicates how many entities overlap at the current position. (c) shows the changing returns of the training process. (d) and (e) show the results of the generalization experiment by changing entities and tasks. (f) is the training curve with worker entities.}
  \label{RBF}
\end{figure*}

%From Figure \ref{elec} we can see that our two-stage method achieves the best performance. Compared to sequential allocation and random allocation methods, our pre-assign method does not fall into local convergence and achieves better allocation results. In the pre-assign module, compared to the fully connected layer used in traditional reinforcement learning algorithms, the improvement by using the proposed TAM module and AMIX module is not significant, but the introduction of these modules can enable the algorithm to have better generalization.

\subsection{Resource-Based Foraging (RBF)}

This experiment will answer questions (2) and (3). This environment is a task assignment environment based on the benchmark environment Level-Based Foraging (LBF, \cite{papoudakis2020benchmarking}) for multi-agent fully collaborative tasks and has undergone some changes to adapt to the settings in this article. The entities possess numerous resources in the RBF environment, but just one resource in the LBF environment, which represents the entities' level. For simplicity, all tasks are still represented by apples, and the differences in task resources are reflected in the size of the apples. The number of entities in RBF is very large, and selecting each entity requires a cost. Apples refresh randomly on the map over time, and the manager needs to control the entity to finish the task, i.e., pick the apples to get rewards. If the task is not completed after a certain period of time, its rewards and requirements will be reduced. Furthermore, if an entity promises to complete a task, it cannot be arranged for other tasks to be completed. \textcolor{black}{In this environment, there are a total of 100 selectable entities, and every 5 time steps, 5-10 apples are randomly generated at any location on the map. The manager needs to select entities from the pool of 100 based on their attributes and positions, determining whether to choose each entity and directing them to solve specific tasks. This can be considered a large-scale task assignment problem, as there are 100 entities to allocate and a considerable number of tasks, requiring sophisticated algorithm design to address the challenge of allocating multiple tasks to a large number of entities.}

We first consider that these entities are item entities whose cost is determined by the size of resources and their distance from tasks. We simplify different tasks as harvesting different apples, each with its own attributes representing the task requirements. A task is considered completed, or an apple successfully harvested, only when the total attributes of the entity at that location exceed those of the apple. Since the entity attributes are manually specified, we train only the manager in this scenario. The objective is to enable the manager to appropriately select entities based on their attributes, corresponding locations, apple positions, and the total entity attributes required for apple harvesting, and then allocate tasks accordingly. The training curve is shown in Figure \ref{RBF}. To verify the generalization ability of algorithms, we first replaced the training entities with another batch of entities we had never seen before. Then we change the attributes of tasks in the environment to see the result. In the task generalization experiment, we randomly generate new apples with different attributes on the map, distributed across various locations, and have the manager utilize the same set of \textcolor{black}{entities} to complete them. This task can be likened to a scenario where a company employs the same group of employees to complete historical tasks and verifies whether they can allocate the employees reasonably based on the requirements of new tasks. The results of the training process, the zero-shot generalization, the few-shot generalization, and the result of training from scratch are shown in Figure \ref{RBF}. We can see that whether it is a change in the number and attributes of \textcolor{black}{entities} or a change in the attributes of tasks, the experimental results of zero-shot generalization of the model are almost consistent with the effect of retraining the model about 700 epochs, and the few-shot training performance of the model, which trains 200 epochs in this environment, is better than retraining it 1000 times. This demonstrates that our model does not remember how it should behave in the current environment but rather learns to assign strategies based on \textcolor{black}{entity} and task attributes. 

We also consider the situation where the \textcolor{black}{entities} are worker entities, and they will demand their costs for being selected to do a task. They will dynamically adjust their quotes based on the attributes of the task, their own attributes, quotes, final returns, and the selected situation. We use the DDPG algorithm to simulate the process of \textcolor{black}{entity} learning to propose demand. We can see from the Figure \ref{RBF} that as the manager uses our algorithms to learn how to distribute, both total and manager benefits increase. And workers gradually increase their quotes and benefits, so the rate of manager revenue increase begins to decrease. As the total income begins to reach the upper limit, the manager's income at this time reaches the upper limit, and workers no longer raise their demands, creating a dynamic balance. In the training process, workers are constantly learning and modifying the quotation, at which time the algorithm can select the appropriate workers to complete the task. This further proves that our algorithm can dynamically give a reasonable distribution result based on the current situation.

\begin{figure*}[t]
    \centering
    \subfigure[]{
    \begin{minipage}[t]{0.35\linewidth}
    \centering
    \includegraphics[height=140pt,width=140pt]{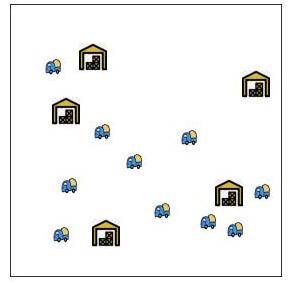}
    \end{minipage}
    }
    \subfigure[]{
    \begin{minipage}[t]{0.6\linewidth}
    \centering
    \includegraphics[height=140pt,width=220pt]{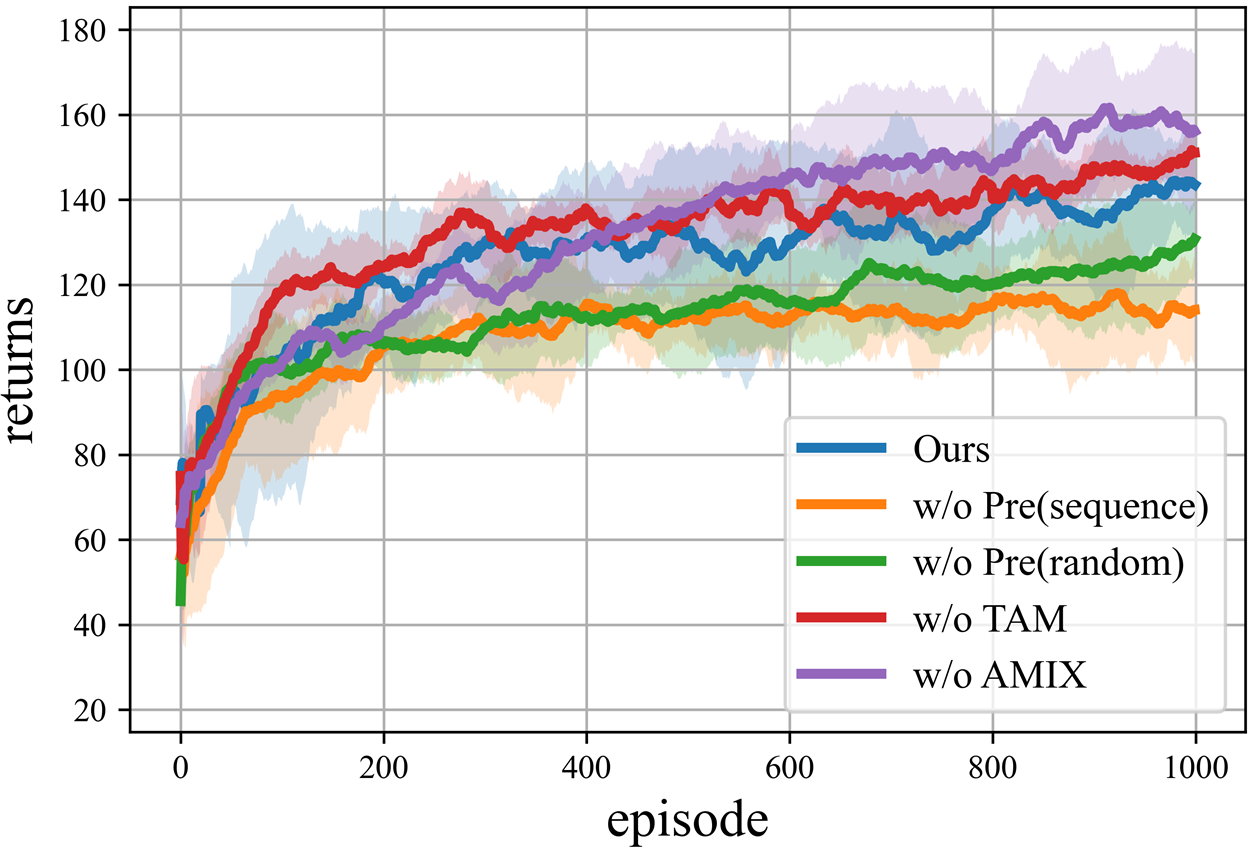}
    \end{minipage}
    }
 \caption{The experiment of Material Transportation.}
  \label{Material}
\end{figure*}

\subsection{Material Transportation}
We use a Material Transportation environment and the problem to be solved is the Dynamic Vehicle Routing Problem (DVRP) problem. There are trucks carrying resources at various parking points on the map, and we need to allocate these trucks reasonably based on the amount of resources required by each city. \textcolor{black}{
We have a total of 50 entities equipped with various materials. At each time step, the environment generates material demands at random positions. The manager needs to dynamically allocate tasks to the agents based on the entities' positions and remaining material attributes.} This environment is different from the previous environments since once an entity completes a task in this environment, the resources that it carries will be cleared, meaning that it cannot be selected to complete other tasks anymore. The results of this experiment are shown in Figure \ref{Material}. From the comparison, it is evident that our algorithm does not achieve the highest performance when compared to the results obtained without the AMIX or the TAM network structure. However, as mentioned earlier, networks lacking these modules are unable to handle variable number entities allocation. Therefore, while our algorithm may not excel in this specific task, it possesses the crucial ability to generalize and adapt to different scenarios.

\subsection{Results}
\textcolor{black}{Here, we compare our proposed reinforcement learning algorithm with three mainstream heuristic algorithms, Genetic Algorithm (\cite{holland1992adaptation}, GA), Particle Swarm Optimization (\cite{kennedy1995particle}, PSO), and Symbiotic Organism Search (\cite{cheng2014symbiotic}, SOS) to address question (4) and demonstrate the effectiveness of our method in dynamic task allocation. As our tasks are dynamically provided, we periodically rerun these heuristic algorithms to adapt to changing task allocations. These algorithms, originally designed for static planning problems, are now triggered for reruns every 10 interactions with the environment, considering the current set of agents and tasks. For these heuristic algorithms, the encoding length is determined by the product of the MDP length and the action space dimension. During interaction with the environment, actions are chosen based on this encoding.
In the Genetic Algorithm, the mutation rate is set to 0.05, meaning each gene in the encoding has a 5\% chance of undergoing a random change. The crossover rate is set to 0.5, with each crossover operation randomly selecting a crossover point and exchanging subsequent genes between parents. The population size is fixed at 100, and the algorithm runs for a total of 1000 generations, which is the same as the RL method. Since each particle represents an allocation, values are constrained between 0 and the maximum number $N+2$. When the integer part equals $k$ ($k \geq 1$), it indicates that the entity is assigned to task $k$; when the integer part is 0 or larger than $N$, it indicates that the entity is not assigned to any task. For the PSO algorithm, we initialize 100 particles, set the inertia weight to 0.5, and both $c_1$ and $c_2$ to 1.5. The total number of iterations is set to 1000. In the SOS algorithm, the number of organisms is also set to 100. The search space spans from 0 to the maximum number plus one, similar to the GA setting. The coefficients during the mutualism and commensalism phases are set as random numbers between -1 and 1.
} 

\begin{figure*}[b]
    \centering
    \subfigure[]{
    \begin{minipage}[t]{0.3\linewidth}
    \centering
    \includegraphics[height=120pt,width=150pt]{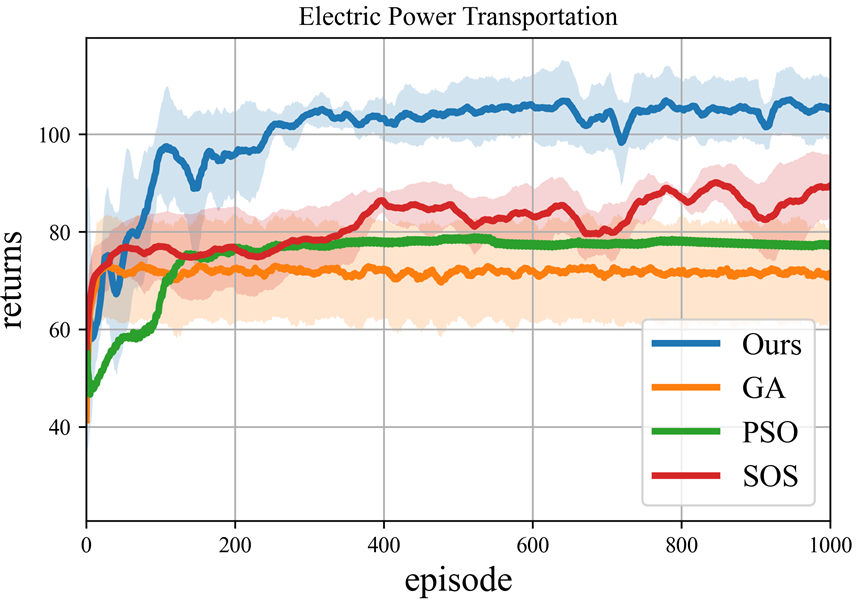}
    \end{minipage}
    }
    \subfigure[]{
    \begin{minipage}[t]{0.3\linewidth}
    \centering
    \includegraphics[height=120pt,width=150pt]{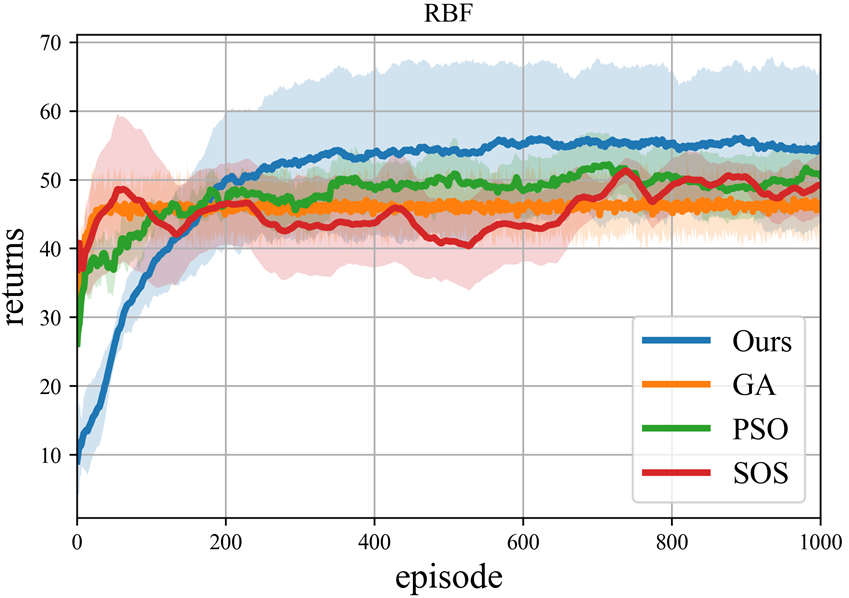}
    \end{minipage}
    }
    \subfigure[]{
    \begin{minipage}[t]{0.3\linewidth}
    \centering
    \includegraphics[height=120pt,width=150pt]{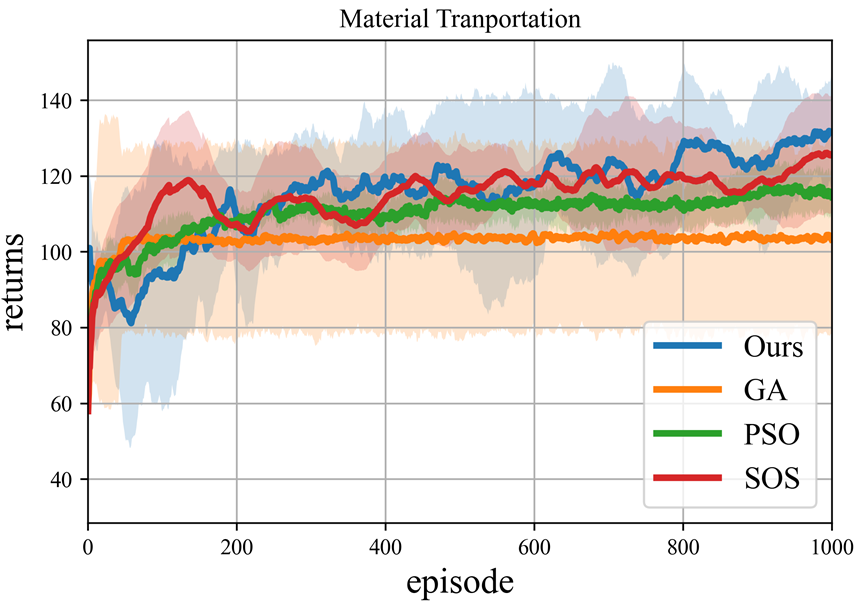}
    \end{minipage}
    }
 \caption{\textcolor{black}{The results of our method, GA, PSO, and SOS methods. We run each algorithm with 5 different seeds and displayed the mean and variance curves. It can be observed that our proposed algorithm achieves better performance compared to heuristic algorithms.}}
  \label{he}
\end{figure*}

The training curves between the number of training steps and the final return are shown in Figure \ref{he}. It can be observed that our proposed two-stage reinforcement learning approach outperforms heuristic algorithms in dynamic task allocation planning. The final returns of the Electric Power Transportation (EPT), Resource-Based Foraging (RBF), and Material Transportation (MT) environments are presented in Table \ref{TABLE}. We conducted a comparative analysis to evaluate the training performance and generalization ability of our proposed method against traditional approaches in these environments. In the entity-generalization task, in the EPT environment, we did not reduce the number of entities but instead changed the attributes of the entities. In the MT environment, we slightly decreased the number of entities from 50 to 40 and also changed the amount of resources carried by each entity. In the RBF environment, we reduced the number of entities from 100 to 50. For the task-generalization scenarios, we altered the probability of occurrence for each task at each time step and replaced the requirements or resources needed for the tasks. The table clearly illustrates that our proposed structure outperforms traditional structures in dynamic task allocation problems. Our algorithm achieved superior performance on most tasks, and the generalization experiments of zero-shot and few-shot learning have demonstrated its adaptability to unseen entities and tasks, resulting in excellent allocation performance.
\begin{table}[t]
    \centering
    \setlength{\tabcolsep}{1mm}{
    \resizebox{\linewidth}{35mm}{
    \begin{tabular}{c c c c c c c c c c}
        \toprule
       \textbf{Env Name}& \textbf{Train$\setminus$Test}& \textbf{Ours} &\textcolor{black}{\textbf{GA}}&\textcolor{black}{\textbf{PSO}} &\textcolor{black}{\textbf{SOS}} & \textbf{W/o Pre(sequence)} & \textbf{W/o Pre(random)}  & \textbf{W/o AMIX} & \textbf{W/o TAM} \\
        \midrule
         \multirow{5}{*}{EPT} & train&\textbf{105.4}$\pm$6.34 
         &\textcolor{black}{72.5$\pm$10.4}&\textcolor{black}{78.6$\pm$4.2} &\textcolor{black}{89.4$\pm$6.5} & 82.4$\pm$8.13 & 89.6$\pm$6.87 & 101.5$\pm$1.44  & 101.0$\pm$1.24\\
        &zero-shot(entity) & 125.5$\pm$0.51&\textcolor{black}{$\setminus$} &\textcolor{black}{$\setminus$}&\textcolor{black}{$\setminus$}  & 92.8$\pm$5.63&101.7$\pm$10.6 & \textbf{126.3}$\pm$1.61 & $\setminus$\\
        & few-shot(entity) &\textbf{130.1}$\pm$2.47&\textcolor{black}{$\setminus$} &\textcolor{black}{$\setminus$} &\textcolor{black}{$\setminus$} & 116.6$\pm$1.63 & 121.3$\pm$1.64 &$\setminus$ & $\setminus$\\
          &zero-shot(task)& 75.1$\pm$0.65&\textcolor{black}{$\setminus$} &\textcolor{black}{$\setminus$}&\textcolor{black}{$\setminus$}  & 64.3$\pm$0.75 & 68.3$\pm$0.83& \textbf{78.3}$\pm$1.03& $\setminus$\\
          &few-shot(task)& \textbf{82.3}$\pm$2.38 &\textcolor{black}{$\setminus$} &\textcolor{black}{$\setminus$}&\textcolor{black}{$\setminus$}  & 73.4$\pm$1.87 &77.4$\pm$2.59 & $\setminus$ & $\setminus$\\
          \hline
          \multirow{5}{*}{RBF}&train& \textbf{54.9}$\pm$11.12&\textcolor{black}{46.2$\pm$3.3} &\textcolor{black}{50.7$\pm$4.0} &\textcolor{black}{48.1$\pm$3.4} & 22.2$\pm$0.93& 46.3$\pm$2.94 &43.4$\pm$17.85  & 36.5$\pm$9.55\\
          &zero-shot(entity)& \textbf{45.7}$\pm$0.67&\textcolor{black}{$\setminus$} &\textcolor{black}{$\setminus$}&\textcolor{black}{$\setminus$}  & 39.6$\pm$0.83 & 35.6$\pm$0.73& 27.4$\pm$0.58 & $\setminus$\\
          &few-shot(entity)& \textbf{50.2}$\pm$1.12&\textcolor{black}{$\setminus$} &\textcolor{black}{$\setminus$}&\textcolor{black}{$\setminus$}  & 46.4$\pm$1.67 & 38.6$\pm$2.43 & $\setminus$ & $\setminus$\\
          &zero-shot(task)&\textbf{62.2}$\pm$1.19&\textcolor{black}{$\setminus$} &\textcolor{black}{$\setminus$}&\textcolor{black}{$\setminus$}  & 26.4$\pm$0.98 &55.1$\pm$1.21 &  36.7$\pm$0.85 & $\setminus$\\
          &few-shot(task)& \textbf{65.9}$\pm$1.64&\textcolor{black}{$\setminus$} &\textcolor{black}{$\setminus$}&\textcolor{black}{$\setminus$}  & 30.3$\pm$2.41 & 57.5$\pm$1.54&  $\setminus$& $\setminus$\\
          \hline
         \multirow{5}{*}{MT} &train& 143.5$\pm$13.9&\textcolor{black}{103.4$\pm$12.5} &\textcolor{black}{114.7$\pm$7.2}&\textcolor{black}{125.7$\pm$7.8} & 114.1$\pm$6.06 & 130.7$\pm$8.79& 151.0$\pm$6.17 & \textbf{156.1}$\pm$18.2\\
          &zero-shot(entity)& 138.6$\pm$3.87&\textcolor{black}{$\setminus$} &\textcolor{black}{$\setminus$}&\textcolor{black}{$\setminus$}  & 98.6$\pm$2.81 & 110.3$\pm$2.75& \textbf{143.6}$\pm$2.54 & $\setminus$\\
          &few-shot(entity)& \textbf{148.0}$\pm$9.56 &\textcolor{black}{$\setminus$} &\textcolor{black}{$\setminus$}&\textcolor{black}{$\setminus$}  & 120.3$\pm$6.12 & 130.3$\pm$4.10&  $\setminus$&$\setminus$ \\
          &zero-shot(task)& 54.3$\pm$1.49&\textcolor{black}{$\setminus$} &\textcolor{black}{$\setminus$}&\textcolor{black}{$\setminus$}  & 44.8$\pm$3.56 &40.4$\pm$2.87 & \textbf{56.3}$\pm$2.23 & $\setminus$\\
          &few-shot(task)& \textbf{60.4}$\pm$3.62&\textcolor{black}{$\setminus$}&\textcolor{black}{$\setminus$}&\textcolor{black}{$\setminus$} & 49.7$\pm$2.65 & 53.6$\pm$4.13& $\setminus$ & $\setminus$\\
         
        \bottomrule 
        \end{tabular}
        }
        
        }
\caption{Analysis of our model. Our proposed network can achieve good performance in the environment and achieve generalization of zero-shot and few-shot.}
\label{TABLE}
\end{table}

\subsection{Analysis}
\textcolor{black}{From the tables and experimental curves, it is clear that our proposed method outperforms in addressing dynamic task allocation problems. From the results presented in Table \ref{TABLE}, it is evident that our proposed method outperforms GA, PSO, and SOS algorithms across all three experimental environments, demonstrating its effectiveness in solving dynamic allocation problems. As shown in Figure \ref{he}, our training curve starts off lower than the heuristic algorithms. However, in the mid-training phase, our method surpasses the heuristics and maintains the highest performance until the end. This may be due to the neural network's need to extract features based on the current attributes of entities and tasks, which requires observing and understanding each dimension and position. This process demands a substantial amount of data to update feature extraction parameters effectively, allowing the model to learn an optimal allocation strategy based on the state of entities and tasks. In the later stages of training, when data volume increases, the reinforcement learning method's learned strategy can provide better real-time task allocation according to the scenario, resulting in a higher performance ceiling. This success can also be attributed to our modeling of the problem as an MDP, which enables real-time capture of changes in environmental tasks and entity attributes. In contrast, heuristic algorithms address static task allocation problems in each iteration and lack the adaptability of reinforcement learning algorithms in handling such dynamic scenarios. Additionally, compared to heuristic algorithms, our method can address generalization issues. When new entities or tasks appear, our algorithm can directly allocate tasks, whereas heuristic algorithms fail to handle generalization due to mismatched dimensions caused by changes in the number of entities or tasks. Furthermore, heuristic algorithms do not take into account the attributes of entities or tasks, so they cannot dynamically adjust their strategies when these attributes change.}

\textcolor{black}{Compared to sequential allocation or random sequential allocation, our proposed pre-allocation method achieves better results. In various environments, our method shows significant improvements, indicating that our pre-allocation approach can find better task allocation solutions. In contrast, task allocation based on natural or random order is more likely to get stuck in local optima, resulting in suboptimal allocations.}

\textcolor{black}{From Table \ref{TABLE}, it's noteworthy that in some experiments, the absence of the AMIX module leads to better results in zero-shot scenarios. This can be attributed to the SHN network's utilization of an attention structure that inherently involves a large number of parameters. The SHN network's output serves as weight parameters for the TAM network. While this approach enhances fine-tuning capabilities across various tasks and scenarios, captures variations in the number of entities, and achieves superior generalization, it also increases the total number of parameters and training complexity compared to directly learning the parameters of the AMIX neural network. Consequently, zero-shot performance may slightly degrade.}

\textcolor{black}{In the RBF environment, employing the AMIX module results in superior performance. However, in other environments, its use leads to slightly inferior results. This variation can be attributed to the specific settings of each environment. For instance, EPT involves fewer entities and simpler tasks. In the generalization scenario, we did not modify the number of entities or total tasks but only altered the resource attributes carried by the entities and the attributes of the tasks. In the RBF environment, we employed 100 entities, whereas in zero-shot and few-shot scenarios, we reduced the number of entities and tasks by 50\%. Since the attention module can better capture the impact of changes in the number of entities on task allocation, our method achieves optimal results in training and various generalization tasks within the RBF environment. It significantly outperforms the AMIX network without the attention module. Conversely, in EPT and MT scenarios, training without the attention module leads to faster convergence. Due to minimal changes in the number of entities, omitting the AMIX module yields better performance. Our proposed method performs better in task environments with large-scale and numerous entities. When there are many entities, the AMIX module can better capture the connections between entities. However, for smaller-scale tasks, the reduced number of entities makes it challenging to capture the relationships between entities and different entity counts. As a result, the proposed module does not show significant improvement during training. Nonetheless, without the AMIX module, the critic cannot accurately estimate task allocation values for a variable number of entities and tasks. It relies solely on the actor module with the attention network to output policies but cannot use the critic network to update those policies. Thus, it is not possible to fine-tune parameters for new tasks.
}

\textcolor{black}{From the perspective of generalization, our proposed method demonstrates excellent generalization without additional training. In Figure \ref{RBF}, directly applying the trained model to new environments yields results comparable to those after approximately 700 steps of retraining. Furthermore, after fine-tuning for 100 steps, it performs better than retraining for 1000 steps, proving the zero-shot and few-shot capabilities of our method. Additionally, compared to sequential and random selection methods, our pre-allocation method also achieves better results in generalization scenarios, further validating the effectiveness and rationality of our approach.}

\section{Conclusion}
In our work, we propose a two-stage approach to solve the task allocation problem. Our approach starts by pre-assigning entities to tasks that they are good at, based on the similarity of each entity and task. We then select from the candidate entities in each task using a sequence model similar to point-net. The proposed TAM and AMIX network architectures can accommodate the changing number of tasks and entities and have the potential to achieve zero-shot and few-shot generalization to new tasks and unseen entities scenarios. Through a variety of experiments, we verify the effectiveness of our proposed two-stage task allocation approach and the validity of our proposed structure. \textcolor{black}{We compared our algorithm with heuristic algorithms in multiple environments and found that our algorithm achieves better results, demonstrating the effectiveness of our approach. Additionally, we conducted generalization experiments by modifying the number of entities and tasks, as well as the associated resources and attributes. This validation confirms that our method can successfully address these challenges and achieve good generalization performance.}

\section{Limitation}
A limitation of our method pertains to the prerequisite knowledge of task and entity attributes for efficient task allocation, which is highly justified in resource scheduling and delivery scenarios. However, in certain contexts, it becomes imperative to make reasonably estimated attributions for tasks and entities. For instance, within an enterprise, quantifying project complexity and required competencies, while conducting quantitative evaluations of employees, becomes indispensable to leverage our algorithm for task allocation. Encouragingly, this limitation can be overcome as certain task characteristics and entity observations are readily obtainable. By employing appropriate methods to extract task and entity features and employing them as attributes, these challenges can be effectively addressed.

% \section{Acknowledgement}
% This work was supported in part by the Science and Technology Innovation 2030-Key Project under Grant 2021ZD0201404. The authors would like to thank members of Tsinghua SIGS Intelligent Computing Lab for their valuable suggestions on the initial version of this manuscript.

\printcredits
\
\bibliographystyle{cas-model2-names}

\bibliography{cas-refs}



\newpage
\appendix

\section{Proof}
\label{APPENDIXA}
The following is a proof of the probability upper bound of the sequential selection method. We define $w\in b_{a_i}$ as the set of entities which only belong to $b_i$, i.e., $w\in b_{a_i}\Leftrightarrow  \forall i,j\ne i, w\in b_i,w\notin b_j$. We denote $r_i=b_i\setminus b_{a_i}$. The number of entities in $b_{a_i}$ and $r_i$ is defined as $n_{a_i}$ and $n_{r_i}$. The summary of $n_{a_i}$ is less than $n-1$ and $n_{r_i}$ is bigger than 1 since $\forall i,w_o\in r_i$. We record the number of entities in $b_{a_j}$ who were mistakenly selected by $T_i$ as $m^i_j$. The probability of $w_o\in T_N$ is

\begin{align*}
&\frac{n_{a_1}}{n_{a_1}+r_{a_1}}\frac{n_{a_2}-m^1_2}{n_{a_2}-m^1_2+r_{a_2}}\frac{n_{a_3}-m^1_3-m^2_3}{n_{a_3}-m^1_3-m^2_3+r_{a_3}}...\frac{n_{a_{N-1}}-\sum_{k = 1}^{N-2}m^k_{N-1} }{n_{a_{N-1}}-\sum_{k = 1}^{N-2}m^k_{N-1}+r_{a_{N-1}}}
\\&\le \frac{n_{a_1}}{n_{a_1}+r_{a_1}}\frac{n_{a_2}}{n_{a_2}+r_{a_2}}\frac{n_{a_3}}{n_{a_3}+r_{a_3}}...\frac{n_{a_{N-1}}}{n_{a_{N-1}}+r_{a_{N-1}}}\;\;\;\;  (\frac{a+c}{b+c}\ge\frac{a}{b}\;\;0<a<b,c\ge0)
\\&\le \frac{n_{a_1}}{n_{a_1}+1}\frac{n_{a_2}}{n_{a_2}+1}\frac{n_{a_3}}{n_{a_3}+1}...\frac{n_{a_{N-1}}}{n_{a_{N-1}}+1}
\\&=\frac{1}{1+\frac1{n_{a_1}}}\frac{1}{1+\frac1{n_{a_2}}}\frac{1}{1+\frac1{n_{a_3}}}...\frac{1}{1+\frac1{n_{a_{N-1}}}}
\\&=e^{-[ln(1+\frac1{n_{a_1}})+ln(1+\frac1{n_{a_2}})+...ln(1+\frac1{n_{a_{N-1}}})]}.
\end{align*}

We consider the function $f(x)=ln(1+\frac{1}{x})$. The second derivative of the function is $\frac{2x+1}{x^2(1+x)^2}$ which is positive when $x>0$. So the function $f(x)=ln(1+\frac{1}{x})$ is a convex function.
By using the Jensen inequality of the convex function, we have 
\begin{align*}
\frac{ln(1+\frac1{n_{a_1}})+ln(1+\frac1{n_{a_2}})+...ln(1+\frac1{n_{a_{N-1}}})}{N-1}>ln(1+\frac{N-1}{\sum_{i=1}^{N-1}n_{a_i}}).
\end{align*}
Using the above inequality, we have

\begin{align*}
&e^{-[ln(1+\frac1{n_{a_1}})+ln(1+\frac1{n_{a_2}})+...ln(1+\frac1{n_{a_{N-1}}})]}
\\&\le e^{-(N-1)ln(1+\frac{N-1}{\sum_{i=1}^{N-1}n_{a_i}})}
\\&\le e^{-(N-1)ln(1+\frac{N-1}{n-1})}
\\&= (\frac{n-1}{N+n-2})^{N-1}.
\end{align*}

The probability of $w_o\in T_N$ using pre-assign method is $\frac{1}{N}$ because we only need to make sure the pre-assign is correct. 

\section{Algorithms}
In this section, we give the pseudocode of the algorithms. Algorithm \ref{ALG1} is the main algorithm of our approach. It shows how we assign tasks based on given tasks and entities. Algorithm \ref{ALG2} corresponds to the pre-assign module mentioned in the main text, which is used to pre-assign entities as the candidates for all tasks. Algorithm \ref{ALG3} is the select module used to select and allocate entities for each task. We used Python and PyTorch framework to implement pseudo code.
\renewcommand{\thealgorithm}{1} %这里用来定义算法1，算法2等
    \begin{algorithm}[h]
        \caption{Two-stage Algorithm} %标题
        \label{ALG1}
        \begin{algorithmic}[1] %每行显示行号，1表示每1行进行显示
            \Require Tasks $T=<T_1,T_2,...,T_m>$, entities $w=<w_1,w_2,...,w_n>$
            \Ensure Allocation $a$  
            \State Initialize all network parameters
            \State Get pre-assign policy function and value function $\pi(.|w)$, $Q(w,.)=\text{Pre-Assign}(T,w)$  
            \State Sample pre-assign action $c~\pi(.|w)$ and generate pre-selected entities for each task $w_T={w_{T_1},w_{T_2},...,w_{T_n}}$ according to $c$
            \State Get task allocation $a=\text{Select}(T,w_T)$
            \State \Return $a$
        \end{algorithmic}
    \end{algorithm}

\renewcommand{\thealgorithm}{2} %这里用来定义算法1，算法2等
    \begin{algorithm}[h]
        \caption{Pre-Assign} %标题
        \label{ALG2}
        \begin{algorithmic}[1] %每行显示行号，1表示每1行进行显示
            \Require Tasks $T=<T_1,T_2,...,T_m>$, entities $w=<w_1,w_2,...,w_n>$
            \Ensure Pre-assign policy function $\pi(.|w)$, value function $Q(w,.)$ 
            \State Initialize all network parameters
            \State Compute entity's policy embedding $\boldsymbol{h}_i\gets f^h(w_i)$, value embedding $\boldsymbol{o}_i\gets f^o(w_i)$, task policy embedding $\boldsymbol{g}\gets f^g(T)$ and task value embedding $\boldsymbol{q}\gets f^q(T_i)$. 
            \State Compute policy $\pi(.|w)\gets \text{SoftMax}(\boldsymbol{h}_i^T\boldsymbol{g}/\sqrt{d})$, allocation value $Q_i(w,.)\gets \boldsymbol{o}_i^T\boldsymbol{q}/\sqrt{d})$
            \State Generate AMIX network parameters $W,b=SHN(w)$
            \State Compute total value estimation $Q(w,.)=\text{AMIX}(Q_1(w,.),Q_2(w,.)...,Q_n(w,.))$
            \State \Return $\pi(.|w)$, $Q(w,.)$ 
        \end{algorithmic}
    \end{algorithm}

\renewcommand{\thealgorithm}{3} %这里用来定义算法1，算法2等
    \begin{algorithm}[h]
        \caption{Select} %标题
        \label{ALG3}
        \begin{algorithmic}[1] %每行显示行号，1表示每1行进行显示
            \Require Tasks $T$, entities $w_T$
            \Ensure Allocation $a$
            \State Initialize all network parameters, $t \gets 0$
            \For{$i=1:m$}
                \State $T_i^t \gets T_i$
                \While{$T_i^t[1:] > \boldsymbol{0}$}
                    \State Compute encoder embedding $d_i \gets f^d(w_{T_i})$, task embedding $e_i^t \gets T_i^t$ 
                    \State Compute value $v\gets f^v(T_i,w_{T_i})$
                    \State Compute policy function $\pi(.|e_i^t) \gets \text{SoftMax}(v^{T} \tanh (W_{1} e_i^t+W_{2} d^i))$
                    \State Sample allocation $u_i^t~\pi(.|e_i^t)$
                    \State Update task $T_i^{t+1} = T_i^t+f^a(u_i^t)$
                \EndWhile
                \State $a_i \gets (u_i^0,u_i^1,...)$
            \EndFor
            \State $a \gets (a_1,a_2,...,a_m)$
            \State \Return $a$ 
        \end{algorithmic}
    \end{algorithm}
    
\section{Experiment details}
Our experiments were performed by using the following hardware and software:
\begin{itemize}
    \item GPUs: NVIDIA GeForce RTX 3090
    \item Python 3.10.8
    \item Numpy 1.23.4
    \item Pytorch 1.13.0
\end{itemize}

\newpage
The hyperparameters of our approach are shown in Table \ref{param}.
\begin{table}[h]
    \centering
\caption{Hyperparameters}
\label{param}
\resizebox{0.9\linewidth}{55mm}{
    \begin{tabular}{l l l}
        \toprule
         \textbf{Name} & \textbf{Description}&\textbf{Value} \\
        \midrule
          lr\_actor& learning rate of all actor network & 1e-4\\ 
          lr\_critic& learning rate of all critic network& 1e-4\\
          \textcolor{black}{lr\_SHR}& \textcolor{black}{learning rate of SHN network}& \textcolor{black}{1e-4}\\
          \textcolor{black}{lr\_AMIX}& \textcolor{black}{learning rate of AMIX network}& \textcolor{black}{1e-4}\\
         optimizer&   type of optimizer & Adam\\
         $\alpha$ & coefficient of entropy loss & 0.05 \\
         $\tau$& soft update rate & 0.005\\ 
          $bs$& batch size & 32\\
          $f$ & activation function & ReLU\\
         $|D|$& maximum size of buffer& 1000\\
       $\gamma$&   discount return & 0.98\\ 
          start $\epsilon$& initial exploration coefficient(manager) & 1\\
         min $\epsilon$& minimum exploration coefficient(manager) & 0.05\\
         $\epsilon$ decay rate & the decay rate of $\epsilon$ when the time step increases& 0.9999\\
         start $\sigma$&   initial exploration coefficient(worker)& 0.5\\
         min $\sigma$& minimum exploration coefficient(worker) & 0.05\\
         $\sigma$ decay rate&  decay rate of $\sigma$ when the time step increases& 0.9999\\
         $h$ & dimensions of all linear hidden layer&64 \\
         $I$& maximum iteration number & 1000\\
         
        \bottomrule 
        \end{tabular}
        }
\end{table}

\end{document}